\documentclass{article}

\usepackage[accepted,nohyperref]{icml2023}

\usepackage{graphicx}
\usepackage{caption}
\usepackage{subcaption}
\usepackage{booktabs} 
\usepackage{nicefrac}
\usepackage{xcolor}
\usepackage{pifont}
\usepackage[inline]{enumitem}
\usepackage{wrapfig}

\usepackage{amssymb}

\usepackage{amsmath,amsthm}
\usepackage{mathtools}
\usepackage{iftex}

\ifxetex
  \usepackage{microtype}
  \usepackage{unicode-math}
  \newcommand{\mathmat}[1]{\mathbfit{#1}}
  \newcommand{\mathvec}[1]{\mathbfit{#1}}
  \newcommand{\mathvecgreek}[1]{\mathbfit{#1}}
  \newcommand{\mathmatgreek}[1]{\mathbfit{#1}}
  
  \newcommand{\boldupright}[1]{\symbfup{#1}}

  \newcommand{\mathrv}[1]{\mathsfit{#1}}
  \newcommand{\mathrvvec}[1]{\mathbfsfit{#1}}
  
  \newcommand{\mathrvvecgreek}[1]{\mathbfsfit{#1}}
  \usepackage{fontspec}
\else
  \usepackage[notext,lcgreekalpha]{stix2}
  \usepackage{times}
  \usepackage{textcomp}

  \newcommand{\mathmat}[1]{\mathbfit{#1}}
  \newcommand{\mathvec}[1]{\mathbfit{#1}}
  \newcommand{\mathvecgreek}[1]{\mathbfit{#1}}
  \newcommand{\mathmatgreek}[1]{\mathbfit{#1}}
  
  \newcommand{\boldupright}[1]{\mathbf{#1}}

  \newcommand{\mathrv}[1]{\mathsfit{#1}}
  \newcommand{\mathrvvec}[1]{\mathbfsfit{#1}}
  
  \newcommand{\mathrvvecgreek}[1]{\mathbfsfit{#1}}
\fi

\usepackage[backgroundcolor=blue!20!white,bordercolor=white]{todonotes}

\usepackage{booktabs,threeparttable,arydshln}
\usepackage{multirow}
\usepackage{nicematrix}

\usepackage[algo2e, ruled]{algorithm2e}

\usepackage{pgfplots}
\pgfkeys{/pgfplots/tuftelike/.style={
  semithick,
  tick style={major tick length=3pt,minor tick length=1.5pt,semithick,black},
  xtick align=outside,
  ytick align=outside,
  separate axis lines,
  axis x line*=bottom,
  axis x line shift=3pt,
  axis y line*=left,
  axis y line shift=3pt,
  }
  }
  
\usepackage{proof-at-the-end}

\usepackage{mdframed}

\declaretheoremstyle[
    bodyfont=\normalfont\itshape,
]{theoremsty}

\mdfdefinestyle {mdleftbarcoloredstyle} {%
    leftline          = true,%
    rightline         = false,%
    topline           = false,%
    bottomline        = false,%
    linewidth         = 2.5pt,%
    backgroundcolor   = gray!10,%
    linecolor         = gray,%
    innertopmargin    = 1.5ex,%
    innerbottommargin = 0.7ex,%
    innerleftmargin   = 1.ex,%
    ntheorem          = false,%
}

\mdfdefinestyle {coloredstyle} {%
    leftline          = false,%
    rightline         = false,%
    topline           = false,%
    bottomline        = false,%
    backgroundcolor   = gray!10,%
    innertopmargin    = 1.ex,%
    innerbottommargin = 0.7ex,%
    innerleftmargin   = 1.ex,%
    userdefinedwidth  = 0.49\textwidth,
}

\declaretheorem[name=Proposition,style=theoremsty, mdframed={style = coloredstyle}]{proposition}

\declaretheorem[name=Theorem,    style=theoremsty, numbered=no]{theorem*}
\declaretheorem[name=Proposition,style=theoremsty, numbered=no]{proposition*}
\declaretheorem[name=Corollary,  style=theoremsty, numbered=no]{corollary*}
\declaretheorem[name=Lemma,      style=theoremsty, numbered=no]{lemma*}

\declaretheoremstyle[
    bodyfont=\normalfont,
]{normalsty}
\declaretheorem[name=Remark,    style=normalsty]{remark}
\declaretheorem[name=Definition,style=normalsty]{definition}
\declaretheorem[name=Assumption,style=normalsty]{assumption}
\declaretheorem[name=Remark,    style=normalsty, numbered=no]{remark*}
\declaretheorem[name=Definition,style=normalsty, numbered=no]{definition*}
\declaretheorem[name=Assumption,style=normalsty, numbered=no]{assumption*}

\newenvironment{proofsketch}{%
  \proof%
  }{\endproof}

\usepackage[many]{tcolorbox}
\tcolorboxenvironment{framedtheorem}{
  colback=blue!5!white,
  boxrule=0pt,
  boxsep=1pt,
  left=2pt,right=2pt,top=2pt,bottom=2pt,
  oversize=2pt,
  sharp corners,
  before skip=\topsep,
  after skip=\topsep,
}

\usepackage{url}
\usepackage[bookmarksopen=true,bookmarks=true,colorlinks=true,backref=page]{hyperref}
\usepackage[noabbrev, capitalise, nameinlink]{cleveref}

\renewcommand*{\backref}[1]{}
\renewcommand*{\backrefalt}[4]{({\footnotesize%
    \ifcase #1 Not cited.%
          \or page~#2%
          \else pages #2%
    \fi%
    })}

\definecolor{linkcolor}{HTML}{005D5D}
\definecolor{citecolor}{HTML}{00539A}
\hypersetup{colorlinks={true},linkcolor=linkcolor,citecolor=citecolor}

\crefname{assumption}{Assumption}{Assumption}

\crefname{framedtheorem}{Theorem}{Theorems}
\crefname{framedproposition}{Proposition}{Propositions}
\crefname{framedlemma}{Lemma}{Lemmas}

\makeatletter
\def\adl@drawiv#1#2#3{%
        \hskip.5\tabcolsep
        \xleaders#3{#2.5\@tempdimb #1{1}#2.5\@tempdimb}%
                #2\z@ plus1fil minus1fil\relax
        \hskip.5\tabcolsep}
\newcommand{\cdashlinelr}[1]{%
  \noalign{\vskip\aboverulesep
           \global\let\@dashdrawstore\adl@draw
           \global\let\adl@draw\adl@drawiv}
  \cdashline{#1}
  \noalign{\global\let\adl@draw\@dashdrawstore
           \vskip\belowrulesep}}
\makeatother

\pgfkeys{/prAtEnd/global custom defaults/.style={
    end,
    restate,
    text proof={Proof},
    text link={\textit{Proof.} See the \hyperref[proof:prAtEnd\pratendcountercurrent]{\textit{full proof}} in~\cref{proof:prAtEnd\pratendcountercurrent}.}
  }
}

\DeclareMathOperator*{\minimize}{minimize}

\DeclareMathOperator*{\argmin}{arg\,min} 

\newcommand*\xbar[1]{%
  \hbox{%
    \vbox{%
      \hrule height 0.6pt 
      \kern0.33ex
      \hbox{%
        \kern-0.1em
        \ensuremath{#1}%
        \kern-0.1em
      }%
    }%
  }%
}

\newcommand{\Esub}[2]{\mathbb{E}_{#1}\left[ #2 \right]}
\newcommand{\V}[1]{\mathbb{V}\left[ #1 \right]}

\newcommand{\DKL}[2]{\mathrm{D}_{\mathrm{KL}}(#1, #2)}

\newcommand{\norm}[1]{{\left\lVert #1 \right\rVert}}
\newcommand{\abs}[1]{{\left| #1 \right|}}

\newcommand{\vg}{\mathvec{g}}

\newcommand{\vm}{\mathvec{m}}

\newcommand{\vs}{\mathvec{s}}
\newcommand{\vt}{\mathvec{t}}
\newcommand{\vu}{\mathvec{u}}

\newcommand{\vw}{\mathvec{w}}
\newcommand{\vx}{\mathvec{x}}
\newcommand{\vy}{\mathvec{y}}
\newcommand{\vz}{\mathvec{z}}

\newcommand{\vlambda}{\mathvecgreek{\lambda}}

\newcommand{\vzeta}{\mathvecgreek{\zeta}}

\newcommand{\rvu}{\mathrv{u}}

\newcommand{\rvx}{\mathrv{x}}

\newcommand{\rvvg}{\mathrvvec{g}}

\newcommand{\rvvu}{\mathrvvec{u}}

\newcommand{\rvvx}{\mathrvvec{x}}

\newcommand{\rvvz}{\mathrvvec{z}}

\newcommand{\rvvzeta}{\mathrvvecgreek{\zeta}}

\newcommand{\mA}{\mathmat{A}}

\newcommand{\mC}{\mathmat{C}}
\newcommand{\mD}{\mathmat{D}}

\newcommand{\mJ}{\mathmat{J}}

\newcommand{\mL}{\mathmat{L}}

\newcommand{\mU}{\mathmat{U}}

\newcommand{\mX}{\mathmat{X}}

\newcommand{\mSigma}{\mathmatgreek{\Sigma}}
\newcommand{\mPhi}{\mathmatgreek{\Phi}}

\newcommand{\inner}[2]{\left\langle #1, #2 \right\rangle}

\usepackage{etoc}
\etocframedstyle[2]{\textbf{\textsc{Table of Contents}}}
\etocsettocdepth{3}

\usepackage{tikz}
\usetikzlibrary{patterns}
\usetikzlibrary{spy}
\usetikzlibrary{arrows.meta}
\usetikzlibrary{pgfplots.groupplots}
\usepgfplotslibrary{fillbetween}
\usepgfplotslibrary{external} 

\pgfplotsset{
    name nodes near coords/.style={
        every node near coord/.append style={
            name=#1-\coordindex,
            alias=#1-last,
        },
    },
    name nodes near coords/.default=coordnode
}

\definecolor{color1}{HTML}{EE5396}
\definecolor{color2}{HTML}{A56EFF}
\definecolor{color3}{HTML}{4589FF}
\definecolor{color4}{HTML}{1192E8}

\newcommand{\lemmaproofoption}{%
  \ifx\debugmode\undefined
    all end%
  \else
    debug%
  \fi
}%

\newcommand{\keylemmaproofoption}{%
  \ifx\debugmode\undefined
    end%
  \else
    debug%
  \fi
}%

\newcommand{\theoremproofoption}{%
  \ifx\debugmode\undefined
  \else
    debug%
  \fi
}%

\icmltitlerunning{Gradient Variance Bounds for Black-Box Variational Inference}

\begin{document}

\twocolumn[
\icmltitle{Practical and Matching Gradient Variance Bounds for \\ Black-Box Variational Bayesian Inference}




\begin{icmlauthorlist}
\icmlauthor{Kyurae Kim}{upenncis}
\icmlauthor{Kaiwen Wu}{upenncis}
\icmlauthor{Jisu Oh}{ncsustat}
\icmlauthor{Jacob R. Gardner}{upenncis}
\end{icmlauthorlist}

\icmlaffiliation{upenncis}{Department of Computer and Information Sciences, University of Pennsylvania, Philadelphia, Pennsylvania, United States}
\icmlaffiliation{ncsustat}{Department of Statistics, North Carolina State University, Raleigh, North Carolina, United States}

\icmlcorrespondingauthor{Kyurae Kim}{kyrkim@seas.upenn.edu}
\icmlcorrespondingauthor{Jacob R. Gardner}{jrgardner@seas.upenn.edu}

\icmlkeywords{Bayesian inference, variational inference, stochastic gradient descent, SGD, VI, gradient variance, reparameterization trick}

\vskip 0.3in
]



\printAffiliationsAndNotice{} 

\begin{abstract}
  Understanding the gradient variance of black-box variational inference (BBVI) is a crucial step for establishing its convergence and developing algorithmic improvements.
  However, existing studies have yet to show that the gradient variance of BBVI satisfies the conditions used to study the convergence of stochastic gradient descent (SGD), the workhorse of BBVI.
  In this work, we show that BBVI satisfies a \textit{matching} bound corresponding to the \(ABC\) condition used in the SGD literature when applied to smooth and quadratically-growing log-likelihoods.
  Our results generalize to nonlinear covariance parameterizations widely used in the practice of BBVI.
  Furthermore, we show that the variance of the mean-field parameterization has provably superior dimensional dependence.
\end{abstract}

\section{Introduction}
Variational inference (VI; \citealt{jordan_introduction_1999,blei_variational_2017,zhang_advances_2019}) algorithms are fast and scalable Bayesian inference methods widely applied in fields of statistics and machine learning.
In particular, black-box VI (BBVI; \citealt{ranganath_black_2014,titsias_doubly_2014}) leverages 
stochastic gradient descent (SGD; \citealt{robbins_stochastic_1951,bottou_online_1999}) for inference of non-conjugate probabilistic models.
With the development of bijectors~\citep{kucukelbir_automatic_2017,dillon_tensorflow_2017,fjelde_bijectors_2020}, most of the methodological advances in BBVI have now been abstracted out through various probabilistic programming frameworks~\citep{carpenter_stan_2017,ge_turing_2018,dillon_tensorflow_2017,bingham_pyro_2019,salvatier_probabilistic_2016}.

Despite the advances of BBVI, little is known about its theoretical properties.
Even when restricted to the location-scale family (\cref{def:family}), it is unknown whether BBVI is guaranteed to converge without having to modify the algorithms used in practice, for example, by enforcing bounded domains, bounded support, bounded gradients, and such.
This theoretical insight is necessary since BBVI methods are known to be less robust~\citep{yao_yes_2018,dhaka_robust_2020,welandawe_robust_2022,dhaka_challenges_2021,domke_provable_2020} compared to other inference methods such as Markov chain Monte Carlo.
Although progress has been made to formalize the theory of BBVI with some generality, the gap between our understanding of BBVI and the convergence guarantees of SGD remains open.
For example,~\citet{domke_provable_2019,domke_provable_2020} provided smoothness and gradient variance guarantees. 
Still, these results do not yet yield a full convergence guarantee and do not extend to \textit{nonlinear} covariance parameterizations used in practice.


In this work, we investigate whether recent progress in relaxing the gradient variance assumptions used in SGD~\citep{tseng_incremental_1998,vaswani_fast_2019,schmidt_fast_2013,bottou_optimization_2018,gower_sgd_2019,gower_stochastic_2021,nguyen_sgd_2018} apply to BBVI. These extensions have led to new insights that the structure of the gradient bounds can have non-trivial interactions with gradient-adaptive SGD algorithms~\citep{zhang_adam_2022}.
For example, when the ``interpolation assumption'' (the gradient noise converges to 0;~\citealt{schmidt_fast_2013,ma_power_2018,vaswani_fast_2019}) does not hold, ADAM~\citep{kingma_adam_2015} provably diverges with certain stepsize combinations~\citep{zhang_adam_2022}.
Until BBVI can be shown to conform to the assumptions used by these recent works, it is unclear how these results relate to BBVI.

While the variance of BBVI gradient estimators has been studied before~\citep{xu_variance_2019,domke_provable_2019,mohamed_monte_2020,fujisawa_multilevel_2021}, the connection with the conditions used in SGD has yet to be established.
As such, we answer the following question:
\vspace{-2ex}%
\begin{quote}
  \textit{Does the gradient variance of BBVI conform to the conditions assumed in convergence guarantees of SGD without modifying the implementations used in practice?}
\end{quote}
\vspace{-2ex}%
The answer is yes!
Assuming the target log joint distribution is smooth and quadratically growing, we show that the gradient variance of BBVI satisfies the \textit{ABC} condition (\cref{assumption:abc}) used by~\citet{polyak_pseudogradient_1973,khaled_better_2023,gower_stochastic_2021}.
Our analysis extends the previous result of \citet{domke_provable_2019} to covariance parameterizations involving nonlinear functions for conditioning the diagonal (see \cref{section:covariance_parameterization}), as commonly done in practice.
Furthermore, we prove that the gradient variance of the mean-field parameterization \citep{peterson_mean_1987,peterson_explorations_1989,hinton_keeping_1993} results in better dimensional dependence compared to full-rank ones.

Overall, our results should act as a key ingredient to obtaining a full convergence guarantees of BBVI, as recently done by~\citet{kim_blackbox_2023}.

Our contributions are summarized as follows:
\begin{itemize}
  \vspace{-2ex}
  \setlength\itemsep{-1.5ex}
  \item[\ding{182}] We provide upper bounds on the gradient variance of BBVI that matches the \textit{ABC condition} (\cref{assumption:abc}) used for analyzing SGD.
    \begin{itemize}[leftmargin=1.5em,]
      \item[\ding{228}] \cref{thm:gradient_upper_bound,thm:gradient_upper_bound_kl} do not require any modification of the algorithms used in practice.
      \item[\ding{228}] \cref{thm:gradient_upper_bound_bounded_entropy} achieves better constants under the stronger \textit{bounded entropy} assumption.
    \end{itemize}
    \item[\ding{183}] Our analysis applies to BBVI parameterizations (\cref{section:covariance_parameterization}) widely used in practice (\cref{table:parameterization_survey}).
    \begin{itemize}[leftmargin=1.5em,]
        \item[\ding{228}] \cref{thm:general_variational_gradient_norm_identity} enables the bounds to cover nonlinear covariance parameterizations.
        \item[\ding{228}] \cref{thm:meanfield_u_identity,remark:meanfield_superiority} shows that the gradient variance of the mean-field parameterization has superior dimensional scaling.
    \end{itemize}
    \item[\ding{184}] We provide a matching lower bound (\cref{thm:gradient_lower_bound}) on the gradient variance, showing that, under the stated assumptions, the ABC condition is the weakest assumption applicable to BBVI.
\end{itemize}
  \vspace{-2ex}


\section{Preliminaries}
{
\paragraph{Notation}
Random variables are denoted in serif, while their realization is in regular font.
(\textit{i.e}, \(x\) is a realization of \(\rvx\), \(\vx\) is a realization of the vector-valued \(\rvvx\).)
\(\norm{\vx}_2 = \){\footnotesize\(\sqrt{\inner{\vx}{\vx}} = \sqrt{\vx^{\top}\vx}\)} denotes the Euclidean norm, while \(\norm{\mA}_{\mathrm{F}} =\) {\footnotesize\(\sqrt{\mathrm{tr}\left(\mA^{\top} \mA\right)}\)} is the Frobenius norm, where {\footnotesize\(\mathrm{tr}\left(\mA\right) = \sum^{d}_{i=1} A_{ii} \)} is the matrix trace.
}%

\vspace{-1ex}
\subsection{Variational Inference}
\vspace{-.5ex}
Variational inference~\citep{peterson_mean_1987,hinton_keeping_1993} is a family of inference algorithms devised to solve the problem
{%
  \setlength{\belowdisplayskip}{1.ex} \setlength{\belowdisplayshortskip}{1.ex}%
  \setlength{\abovedisplayskip}{1.ex} \setlength{\abovedisplayshortskip}{1.ex}%
\begin{align}
  \minimize_{\vlambda \in \mathbb{R}^p} \; \DKL{q_{\psi,\vlambda}}{\pi}, \label{eq:kl}
\end{align}
}%
where \(q_{\psi,\vlambda}\) is called the ``variational approximation'', while \(\pi\) is a distribution of interest, and \(D_{\text{KL}}\) is the (exclusive) Kullback-Leibler (KL) divergence.

For Bayesian inference, \(\pi\) is the posterior distribution
{%
  \setlength{\belowdisplayskip}{1.ex} \setlength{\belowdisplayshortskip}{1.ex}%
  \setlength{\abovedisplayskip}{1.ex} \setlength{\abovedisplayshortskip}{1.ex}%
\begin{align*}
  \pi\left(\vz\right)
  \propto 
  \ell\left(\vx \mid \vz \right) p\left(\vz\right)
  =
  \ell\left(\vx, \vz \right),
\end{align*}
}%
where \(\ell\left(\vx \mid \vz \right)\) is the likelihood, and \(p\left(\vz\right)\) is the prior.
In practice, one only has access to the likelihood and the prior.
Thus,~\cref{eq:kl} cannot be directly solved.
Instead, we can minimize the negative \textit{evidence lower bound} (ELBO; \citealt{jordan_introduction_1999}) function \(F\left(\vlambda\right)\).

\vspace{-1.5ex}
\paragraph{Evidence Lower Bound}
More formally, we solve
{%
\setlength{\belowdisplayskip}{1.ex} \setlength{\belowdisplayshortskip}{1.ex}
\setlength{\abovedisplayskip}{1.ex} \setlength{\abovedisplayshortskip}{1.ex}
\[
  \minimize_{\vlambda \in \mathbb{R}^p} \; F\left(\vlambda\right),
\]
}%
where \(F\) is defined as
{%
\setlength{\belowdisplayskip}{1.ex} \setlength{\belowdisplayshortskip}{1.ex}
\setlength{\abovedisplayskip}{1.ex} \setlength{\abovedisplayshortskip}{1.ex}
\begin{align}
  F\left(\vlambda\right) 
  &\triangleq
  -\Esub{\rvvz \sim q_{\psi,\vlambda}}{ \log \ell\left(\vx, \rvvz\right) } - \mathrm{H}\left(q_{\psi,\vlambda}\right),
  \label{eq:elbo_H_form}
  \\
  &=
  -\Esub{\rvvz \sim q_{\psi,\vlambda}}{ \log \ell\left(\vx|\rvvz\right) } + \DKL{q_{\psi,\vlambda}}{p},
  \label{eq:elbo_kl_form}
\end{align}
}%
\begin{center}
  \vspace{-1.5ex}
  {\begingroup
    \setlength\tabcolsep{1.5pt} 
  \begin{tabular}{ll}
    \(\rvvz\) & is the latent (random) variable, \\
    \(q_{\psi,\vlambda}\) & is the variational distribution, \\
    \(\psi\) & is a bijector (support transformation), and  \\
    \(\mathrm{H}\)    & is the differential entropy.
  \end{tabular}
  \endgroup}
\end{center}

The bijector \(\psi\)~\citep{dillon_tensorflow_2017,fjelde_bijectors_2020,leger_parametrization_2023} is a differentiable bijective map that is used to de-constrain the support of constrained random variables.
For example, when \(z\) is expected to follow a gamma distribution, using \(\eta = \psi\left(z\right)\) with \(\psi\left(z\right) = \log z\) lets us work with \(\eta\), which can be any real number, unlike \(z\).
The use of \(\psi^{-1}\) corresponds to the automatic differentiation VI formulation (ADVI;~\citealt{kucukelbir_automatic_2017}), which is now widespread.

\vspace{-2.5ex}
\subsection{Variational Family}
\vspace{-.5ex}
In this work, we specifically consider the location-scale variational family with a standardized base distribution.

\begin{definition}[\textbf{Reparameterization Function}]\label{def:reparam}
  An affine mapping \(\vt_{\vlambda} : \mathbb{R}^d \rightarrow \mathbb{R}^d\) defined as
{
\setlength{\belowdisplayskip}{1.ex} \setlength{\belowdisplayshortskip}{1.ex}
\setlength{\abovedisplayskip}{1.ex} \setlength{\abovedisplayshortskip}{1.ex}
  \begin{align*}
    &\vt_{\vlambda}\left(\vu\right) \triangleq \mC \vu + \vm
  \end{align*}
}%
  with \(\vlambda\) containing the parameters for forming the location \(\vm \in \mathbb{R}^d\) and scale \(\mC = \mC\left(\vlambda\right) \in \mathbb{R}^{d \times d}\) is called the (location-scale) \textit{reparameterization function}.
\end{definition}

\begin{definition}[\textbf{Location-Scale Family}]\label{def:family}
  Let \(\varphi\) be some \(d\)-dimensional distribution.
  Then, \(q_{\vlambda}\) such that
{%
\setlength{\belowdisplayskip}{1.5ex} \setlength{\belowdisplayshortskip}{1.5ex}%
\setlength{\abovedisplayskip}{1.5ex} \setlength{\abovedisplayshortskip}{1.5ex}%
  \begin{alignat*}{2}
    \rvvzeta \sim q_{\vlambda}  \quad\Leftrightarrow\quad &\rvvzeta \stackrel{d}{=} \vt_{\vlambda}\left(\rvvu\right); \quad \rvvu \sim  \varphi
  \end{alignat*}
  }%
  is said to be a member of the location-scale family indexed by the base distribution \(\varphi\) and parameter \(\vlambda\).
\end{definition}

This family includes commonly used variational families, such as the mean-field Gaussian, full-rank Gaussian, Student-T, and other elliptical distributions.
\begin{remark}[\textbf{Entropy of Location-Scale Distributions}]\label{thm:location_scale_entropy}
  The differential entropy of a location-scale family distribution (\cref{def:family}) is 
{
\setlength{\belowdisplayskip}{1.ex} \setlength{\belowdisplayshortskip}{1.ex}
\setlength{\abovedisplayskip}{1.ex} \setlength{\abovedisplayshortskip}{1.ex}
  \[
    \mathrm{H}\left(q_{\vlambda}\right) = \mathrm{H}\left(\varphi\right) + \log \abs{ \mC }.
  \]
}
\end{remark}

\begin{definition}[\textbf{ADVI Family}; \citealt{kucukelbir_automatic_2017}]\label{def:advi}
  Let \(q_{\vlambda}\) be some \(d\)-dimensional distribution.
  Then, \(q_{\psi,\vlambda}\) such that
{
\setlength{\belowdisplayskip}{1.ex} \setlength{\belowdisplayshortskip}{1.ex}
\setlength{\abovedisplayskip}{1.ex} \setlength{\abovedisplayshortskip}{1.ex}
  \begin{alignat*}{2}
    \rvvz \sim q_{\psi,\vlambda}  \quad\Leftrightarrow\quad &\rvvz \stackrel{d}{=} \psi^{-1}\left(\rvvzeta\right); \quad \rvvzeta \sim q_{\vlambda}
  \end{alignat*}
}%
  is said to be a member of the ADVI family with the base distribution \(q_{\vlambda}\) parameterized with \(\vlambda\).
\end{definition}

We impose assumptions on the base distribution \(\varphi\).
\begin{assumption}[\textbf{Base Distribution}]\label{assumption:symmetric_standard}
  \(\varphi\) is a \(d\)-dimensional distribution such that \(\rvvu \sim \varphi\) and \(\rvvu = \left(\rvu_1, \ldots, \rvu_d \right)\) with indepedently and identically distributed components.
  Furthermore, \(\varphi\) is
  \begin{enumerate*}[label=\textbf{(\roman*)}]
      \item symmetric and standardized such that \(\mathbb{E}\rvu_i = 0\), \(\mathbb{E}\rvu_i^2 = 1\), \(\mathbb{E}\rvu_i^3 = 0\), and 
      \item has finite kurtosis \(\mathbb{E}\rvu_i^4 = \kappa < \infty\).
  \end{enumerate*}
\end{assumption}
These assumptions are already satisfied in practice by, for example, generating \(\rvu_i\) from a univariate normal or Student-T with \(\nu > 4\) degrees of freedom.

\subsection{Reparameterization Trick}
\vspace{-.5ex}
When restricted to location scale families~(\cref{def:family,def:advi}), we can invoke Change-of-Variable, or more commonly known as the ``reparameterization trick,'' such that
{%
\begin{align*}
  \mathbb{E}_{\rvvz \sim q_{\psi,\vlambda}} \log\ell\left(\vx, \rvvz\right)
  &=
  \mathbb{E}_{\rvvzeta \sim q_{\vlambda}} \log\ell\left(\vx, \psi^{-1}\left( \rvvzeta\right)\right)
  \\
  &=
  \mathbb{E}_{\rvvu \sim \varphi} \log\ell\left(\vx, \psi^{-1}\left(\vt_{\vlambda}\left(\rvvu\right)\right)\right)
\end{align*}
}%
through the Law of the Unconcious Statistician.
Differentiating this results in the \textit{reparameterization} or \textit{path} gradient, which often achieves lower variance than alternatives~\citep{xu_variance_2019,mohamed_monte_2020a}.

\vspace{-.5ex}
\paragraph{Objective Function}
For generality, we represent our objective as a composite infinite sum problem:
\begin{definition}[\textbf{Composite Infinite Sum}]\label{def:generic_elbo}
\begin{equation*}
   F\left(\vlambda\right)
   =
   \mathbb{E}_{\rvvu \sim \varphi} f\left(\vt_{\vlambda}\left(\rvvu\right)\right)
   +
   h\left(\vlambda\right),
   \label{eq:F}
\end{equation*}
where \((\vlambda, \rvvu) \mapsto f \circ \vt_{\vlambda} : \mathbb{R}^p \times \mathbb{R}^d \rightarrow \mathbb{R}\) is some bivariate stochastic function of \(\vlambda\) and the ``noise source'' \(\rvvu\), while \(h\) is a deterministic regularization term.
\end{definition}
By appropriately defining \(f\) and \(h\), we retrieve the two most common formulations of the ELBO in \cref{eq:elbo_H_form} and \cref{eq:elbo_kl_form} respectively:
\begin{definition}[\textbf{ELBO Entropy-Regularized Form}]\label{def:entropy_form}
\begin{align*}
  f_{\mathrm{H}}\left(\vzeta\right) 
  &= -\log \underbrace{\ell\left( \vx, \psi^{-1}\left( \vzeta \right) \right)}_{\text{Joint Likelihood}} - \log \abs{ \mJ_{\psi^{-1}}\left(\vzeta\right) }  &
  \\
  h_{\mathrm{H}}\left(\vlambda\right) &= - \mathrm{H}\left(q_{\vlambda}\right).
\end{align*}
\end{definition}
\begin{definition}[\textbf{ELBO KL-Regularized Form}]\label{def:kl_form}
\begin{align*}
  f_{\text{KL}}\left(\vzeta\right) 
  &= - \log \underbrace{\ell\left(\vx \mid \psi^{-1}\left( \vzeta \right)  \right)}_{\text{Likelihood}} - \log \abs{ \mJ_{\psi^{-1}}\left(\vzeta\right) } &
  \\
  h_{\text{KL}}\left(\vlambda\right) &= \DKL{q_{\vlambda}}{p}.
\end{align*}
\vspace{-5ex}
\end{definition}%
Here, \(\mJ_{\psi^{-1}}\) is the Jacobian of the bijector.
Since \(\DKL{q_{\vlambda}}{p}\) is seldomly available in tractable form, the entropy-regularized form is the most widely used, while the KL regularized is common for Gaussian processes and variational autoencoders.

\vspace{-1.0ex}
\paragraph{Gradient Estimator}
We denote the \(M\)-sample estimator of the gradient of \(F\) as
\begin{align}
  \rvvg_{M}\left(\vlambda\right) &\triangleq \frac{1}{M} \, \sum^{M}_{m=1} \rvvg_m\left(\vlambda\right), \;\text{where}\; \label{eq:def_gradient_M_est} \\
  \rvvg_m\left(\vlambda\right)   &\triangleq \nabla_{\vlambda} f\left(\vt_{\vlambda}\left(\rvvu_m\right)\right) + \nabla h(\vlambda); \quad \rvvu_m \sim \varphi. \label{eq:def_gradient_m_est}
\end{align}
We will occasionally drop \(\vlambda\) for clarity.

\vspace{-.5ex}
\subsection{Gradient Variance Assumptions in \linebreak Stochastic Gradient Descent}
\vspace{-.5ex}
\paragraph{Gradient Variance Assumptions in SGD}
For a while, most convergence proofs in SGD have relied on the ``bounded variance'' assumption.
That is, for a gradient estimator \(\rvvg\), \(\mathbb{E} \norm{\rvvg}^2_2 \leq G\) for some finite constant \(G\).
This assumption is problematic because
\begin{enumerate*}
  \item[\ding{182}] these types of global constants result in loose bounds, 
  \item[\ding{183}] and it directly contradicts the strong-convexity assumption~\citep{nguyen_sgd_2018}.
\end{enumerate*}
Thus, retrieving previously known SGD convergence rates under weaker assumptions has been an important research direction~\citep{tseng_incremental_1998,vaswani_fast_2019,schmidt_fast_2013,bottou_optimization_2018,gower_sgd_2019,gower_stochastic_2021,nguyen_sgd_2018}.

\vspace{-.5ex}
\paragraph{ABC Condition}\label{section:abc}
In this work, we focus on the recently rediscovered \textit{expected smoothness}, or \textit{ABC}, condition~\citep{polyak_pseudogradient_1973,gower_stochastic_2021}.
\begin{assumption}[\textbf{Expected Smoothness; \(ABC\)}]\label{assumption:abc}
  \(\rvvg\) is said to satisfy the expected smoothness condition if 
  \begin{align*}
    \mathbb{E}\norm{\rvvg_M\left(\vlambda\right)}_2^2
    \leq
    2 \, A \left( F\left(\vlambda\right) - F^* \right)
    + B \, \norm{ \nabla F\left(\vlambda\right) }_2^2 + C.
  \end{align*}
  for some finite \(A, B, C \geq 0\), where \(F^* = \inf_{\vlambda \in \mathrm{R}^p} F\left(\vlambda\right) \).
\end{assumption}
As shown by \citet{khaled_better_2023}, this condition is not only strictly weaker than many of the previously used assumptions but also \textit{generalizes} them by retrieving known convergence rates when tweaking the constants.%

\begin{wraptable}[20]{r}[\dimexpr\columnwidth+\columnsep\relax]{0.65\textwidth}
  \centering
\caption{Survey of Parameterizations Used in Black-Box Variational Inference}\label{table:parameterization_survey}
\vspace{-2ex}
\hspace*{-1.5em}
  {\footnotesize
\setlength{\tabcolsep}{3pt}
\renewcommand{\arraystretch}{1.0}
\begin{threeparttable}
\begin{tabular}{lllcc}
    \toprule
    \multicolumn{1}{c}{\textbf{Framework}} & \multicolumn{1}{c}{\textbf{Version}} & \multicolumn{1}{c}{\textbf{Parameterizations}} & \multicolumn{1}{c}{\textbf{Conditioner}} & \multicolumn{1}{c}{\textbf{Code}} \\ \midrule
    \multirow{1}{*}{ \textsc{Turing}~\citep{ge_turing_2018} } & \multirow{1}{*}{\texttt{v0.23.2}}
    & {Nonlinear} Mean-field & softplus &
    {\footnotesize{\href{https://github.com/TuringLang/Turing.jl/blob/f9353f0f287504a6fa02809cb2e32cc5173b58ae/src/variational/advi.jl#L102}{link}}}
    \\
    \arrayrulecolor{gray!40}\cmidrule{1-5}
    %
    \multirow{2}{*}{ \textsc{Stan}~\citep{carpenter_stan_2017} } & \multirow{2}{*}{\texttt{v2.31.0}}
    & {Nonlinear} Mean-field & exp &
    {\footnotesize\href{https://github.com/stan-dev/stan/blob/58180bc999f5c65d7e8e1af8ad977ad2fc2bc67f/src/stan/variational/families/normal_meanfield.hpp#L28}{link}}
    \\\cmidrule{3-5}
    & & {Linear} Cholesky & &
    {\footnotesize{\href{https://github.com/stan-dev/stan/blob/58180bc999f5c65d7e8e1af8ad977ad2fc2bc67f/src/stan/variational/families/normal_fullrank.hpp#L343}{link}}}
    \\\cmidrule{1-5} 
    \multirow{2}{*}{ \textsc{Pyro}~\citep{bingham_pyro_2019} } & \multirow{2}{*}{\texttt{v0.10.1}}
    & {Nonlinear} Mean-field & softplus & 
    {\footnotesize\href{https://github.com/pyro-ppl/numpyro/blob/aefadc1b78da3b0671fd053d11817d426ef68c8b/numpyro/infer/autoguide.py#L244}{link}}
    \\\cmidrule{3-5}
    & & {Linear} Cholesky\tnote{1} &  &
    {\footnotesize{\href{https://github.com/pyro-ppl/numpyro/blob/aefadc1b78da3b0671fd053d11817d426ef68c8b/numpyro/infer/autoguide.py#L1515}{link}}}
    \\\cmidrule{1-5} 
    \multirow{2}{*}{\textsc{PyMC3}~\citep{salvatier_probabilistic_2016}} & \multirow{2}{*}{\texttt{v5.0.1}}
    & {Nonlinear} Mean-field & softplus &
    {\footnotesize{\href{https://github.com/pymc-devs/pymc/blob/f96594bb215b44197615c695130c9d60e1bf9601/pymc/variational/approximations.py#L61}{link}}}
    \\\cmidrule{3-5}
    & & {Nonlinear} Cholesky & softplus &
    {\footnotesize{\href{https://github.com/pymc-devs/pymc/blob/f96594bb215b44197615c695130c9d60e1bf9601/pymc/variational/approximations.py#L142}{link}}}  
    \\\cmidrule{1-5} 
    \multirow{2}{*}{\textsc{GPyTorch}~\citep{gardner_gpytorch_2018}} & \multirow{2}{*}{\texttt{v1.9.0}}
    & {Linear} Cholesky   &  &
    {\footnotesize{\href{https://github.com/cornellius-gp/gpytorch/blob/071314acca8eb7f4a61afc73e53feb05e1c8cf69/gpytorch/variational/cholesky_variational_distribution.py#L37}{link}}}
    \\\cmidrule{3-5}
    & & {Linear} Mean-field & &
    {\footnotesize{\href{https://github.com/cornellius-gp/gpytorch/blob/071314acca8eb7f4a61afc73e53feb05e1c8cf69/gpytorch/variational/mean_field_variational_distribution.py#L42}{link}}}
    \\
    \arrayrulecolor{black}\bottomrule
\end{tabular}
\begin{tablenotes}
\item[1] Numpyro also provides a low-rank Cholesky parameterization, which is non-linearly conditioned. But the full-rank Cholesky is linear.
\item[*] Tensorflow probability~\citep{dillon_tensorflow_2017} wasn't included as it doesn't provide a fully pre-configured variational family (although \href{https://www.tensorflow.org/probability/api_docs/python/tfp/experimental/vi/build_affine_surrogate_posterior}{\texttt{tfp.experimental.vi.build\_*\_posterior}} exists, the parameterization is user-supplied).
\end{tablenotes}
\end{threeparttable}
  }%
  \vspace{-3ex}
\end{wraptable}

%
%
With the \(ABC\) condition, for nonconvex \(L\)-smooth functions, under a ``appropriately chosen'' stepsize (otherwise the bound may blow-up as explained by~\citeauthor{khaled_better_2023}) of \(\gamma \leq \nicefrac{1}{L B}\), SGD converges to a \(\mathcal{O}\left(L C \gamma \right)\) neighborhood in a \(\mathcal{O}\left(\nicefrac{{\left(1 + L \gamma^2 A\right)}^T}{\left(\gamma T\right)}\right)\) rate.
Minor variants of the \textit{ABC} condition have also been used to prove convergence of SGD for quasar convex functions~\citet{gower_sgd_2021}, stochastic heavy-ball/momentum methods~\citet{liu_almost_2022}, and stochastic proximal methods~\citep{li_unified_2022}.
Given the influx of results based on the \textit{ABC} condition, connecting with it would significantly broaden our theoretical understanding of BBVI.

\vspace{-1.0ex}
\subsection{Covariance Parameterizations}\label{section:covariance_parameterization}
\vspace{-.5ex}

When using the location-scale family (\cref{def:family}), the scale matrix \(\mC\) can be parameterized in different ways.
Any parameterization that results in a positive definite covariance \(\mC\mC^{\top} \in \mathbb{S}_{++}^{d}\) is valid.
We consider multiple parameterizations as the choice can result in different theoretical properties.
A brief survey on the use of different parameterizations is shown in~ \cref{table:parameterization_survey}.

\vspace{-2ex}
\paragraph{Linear Parameterization}
The previous results by~\citet{domke_provable_2019} considered the matrix square root parameterization, which is linear with respect to the variational parameters.
{
\setlength{\belowdisplayskip}{.5ex} \setlength{\belowdisplayshortskip}{.5ex}
\setlength{\abovedisplayskip}{.5ex} \setlength{\abovedisplayshortskip}{.5ex}
\begin{definition}[\textbf{Matrix Square Root}]\label{def:squareroot}
  {
    \[
      \mC\left(\vlambda\right) = \mC,
    \]
    where \(\mC \in \mathbb{R}^{d \times d}\) is a matrix, \(\vlambda_{\mC} = \mathrm{vec}\left(\mC\right) \in \mathbb{R}^{d^2}\) such that \(\vlambda = \left(\vm, \vlambda_{\mC}\right)\).
  }
\end{definition}
}\vspace{-1ex}%
Note that \(\mC\) is not constrained to be symmetric so this is not a matrix square root in a narrow sense.
Also, this parameterization does not guarantee \(\mC\mC^{\top}\) to be positive definite (only positive semidefinite), which occasionally results in the entropy term \(h_{\mathrm{H}}\) blowing up~\citep{domke_provable_2020}.
\citeauthor{domke_provable_2020} proposed to fix this by using proximal operators.

\vspace{-2.0ex}
\paragraph{Nonlinear Parameterizations}
In practice, optimization is preferably done in unconstrained \(\mathbb{R}^p\), which then positive definiteness can be ensured by explicitly mapping the diagonal elements to positive numbers.
We denote this by the \textit{diagonal conditioner} \(\phi\). (See 
\cref{table:parameterization_survey} for a brief survey on their use).
The following two parameterizations are commonly used, where \(\mD = \mathrm{diag}\left(\phi\left(\vs\right)\right) \in \mathbb{R}^{d \times d}\) denotes a diagonal matrix such that \(D_{ii} = \phi\left(s_i\right) > 0\).

\vfill

\vspace*{19\baselineskip}

{
\setlength{\belowdisplayskip}{1.ex} \setlength{\belowdisplayshortskip}{1.ex}
\setlength{\abovedisplayskip}{1.ex} \setlength{\abovedisplayshortskip}{1.ex}
\begin{definition}[\textbf{Mean-Field}]\label{def:meanfield}
  \begin{align*}
    \mC\left(\vlambda, \phi\right) &= \mathrm{diag}\left(\phi\left(\vs\right)\right),
  \end{align*}
  where \(\vs \in \mathbb{R}^{d}\) and \(\vlambda = \left(\vm, \vs\right)\).
\end{definition}

\begin{definition}[\textbf{Cholesky}]\label{def:fullrank}
  \begin{align*}
    \mC\left(\vlambda, \phi\right) &= \mathrm{diag}\left(\phi\left(\vs\right)\right) + \mL,
  \end{align*}
  where \(\vs \in \mathbb{R}^{d}\), \(\mL \in \mathbb{R}^{d \times d}\) is a strictly lower triangular matrix, \(\vlambda_{\mL} = \mathrm{vec}\left(\mL\right) \in \mathbb{R}^{\left(d + 1\right) \, d / 2}\) such that \(\vlambda = \left(\vm, \vs, \vlambda_{\mL}\right)\).
  The special case of \(\phi\left(x\right) = x\) is called the ``linear Cholesky'' parameterization.
\end{definition}
}%
\vspace{-2.5ex}
\paragraph{Diagonal conditioner}
For the diagonal conditioner, the softplus function \(\phi\left(x\right) = \mathrm{softplus}(x) \triangleq \log(1 + e^x)\)~\citep{dugas_incorporating_2000} or the exponential function \(\phi\left(x\right) = e^x\) is commonly used.
While using these nonlinear functions significantly complicates the analysis, assuming \(\phi\) to be 1-Lipschitz retrieves practical guarantees.

\vspace{0.5ex}
\begin{assumption}[\textbf{Lipschitz Diagonal Conditioner}]\label{assumption:phi_lipschitz}
  The diagonal conditioner \(\phi\) is 1-Lipschitz continuous.
\end{assumption}
\vspace{0.5ex}

\begin{remark}
  The softplus function is 1-Lipschitz.
\end{remark}



\section{Main Results}
\vspace{-.5ex}




\subsection{Key Lemmas}
The main challenge in studying BBVI is that the gradient of the composed function \(\nabla_{\vlambda} f \left( \vt_{\vlambda} \left( \vu \right) \right)\) is different from \(\nabla f\).
For the matrix square root parameterization, \citet{domke_provable_2019} established the connection through Lemma 1 (restated as \cref{thm:variational_gradient_norm_identity} in \cref{section:external_lemmas}).
We generalize this result to nonlinear parameterizations:

\begin{theoremEnd}[\keylemmaproofoption,category=upperboundkeylemmagradientnormidentity]{lemma}\label{thm:general_variational_gradient_norm_identity}
  Let \(\vt_{\vlambda}: \mathbb{R}^d \rightarrow \mathbb{R}^d\) be a location-scale reparameterization function (\cref{def:reparam}) with some differentiable function \(f : \mathbb{R}^d \rightarrow \mathbb{R} \).
  Then, for \(\vg_{f} \triangleq \nabla f\left( \vt_{\vlambda}\left(\vu\right) \right)\), 
  \begin{enumerate}[label=(\roman*)]
    \vspace{-2ex}
    \setlength\itemsep{-1ex}
    \item Mean-Field
      \vspace{-1ex}
    {\small
    \setlength{\belowdisplayskip}{1ex} \setlength{\belowdisplayshortskip}{1ex}
    \setlength{\abovedisplayskip}{1ex} \setlength{\abovedisplayshortskip}{1ex}
    \begin{alignat*}{2}
      \hspace{-3em}
      \norm{ \nabla_{\vlambda} f\left( \vt_{\vlambda}\left(\vu\right) \right) }_2^2
      = 
      {\lVert \vg_{f} \rVert}_2^2
      +
      \vg_{f}^{\top}
      \mU \mPhi
      \vg_{f},
      \qquad\qquad
    \end{alignat*}
  }

    \item Cholesky
      {\small%
    {
    \setlength{\belowdisplayskip}{1ex} \setlength{\belowdisplayshortskip}{1ex}
    \setlength{\abovedisplayskip}{1ex} \setlength{\abovedisplayshortskip}{1ex}
    \begin{alignat*}{2}
      \norm{ \nabla_{\vlambda} f\left( \vt_{\vlambda}\left(\vu\right) \right) }_2^2
      &=
      {\lVert \vg_{f} \rVert}_2^2 + \vg_{f}^{\top} \mSigma \vg_{f}
      +
      \vg_{f}^{\top}
      \mU
      \left(
      \mPhi
      - 
      \boldupright{I}
      \right)
      \vg_{f},
    \end{alignat*}
      }%
  }
  \end{enumerate}
  where \(\mU,\mPhi,\mSigma\) are diagonal matrices, which the diagonals are defined as 
  {
    \setlength{\belowdisplayskip}{.5ex} \setlength{\belowdisplayshortskip}{.5ex}
    \setlength{\abovedisplayskip}{.5ex} \setlength{\abovedisplayshortskip}{.5ex}
    \[
    U_{ii} = u_i^2,\quad
    \Phi_{ii} = {\phi^{\prime}\left(s_i\right)}^2,\quad
    \Sigma_{ii} = {\textstyle\sum^{i}_{j=1}} u_j^2,
    \]
  }
  and \(\phi\) is a diagonal conditioner for the scale matrix.
\end{theoremEnd}
\vspace{-2ex}
\begin{proofEnd}
  The proof starts by applying the Chain Rule and then computing the quadratic norm of the gradient as
  \begin{alignat}{2}
    &\norm{\nabla_{\vlambda} f\left( \vt_{\vlambda}\left(\vu\right) \right) }_2^2
    \nonumber
    \\
    &\;= 
    {\left(
      \frac{
        \partial \vt_{\vlambda}\left(\vu\right)
      }{
        \partial \vlambda
      }
      \nabla f\left( \vt_{\vlambda}\left(\vu\right) \right)
    \right)}^{\top}
    \frac{
      \partial \vt_{\vlambda}\left(\vu\right)
    }{
      \partial \vlambda
    }
    \nabla f\left( \vt_{\vlambda}\left(\vu\right) \right)
    \nonumber
    \\
    &\;=
    {\nabla f^{\top}\left( \vt_{\vlambda}\left(\vu\right) \right)}
    {\left(
      \frac{
        \partial \vt_{\vlambda}\left(\vu\right)
      }{
        \partial \vlambda
      }
    \right)}^{\top}
    \frac{
      \partial \vt_{\vlambda}\left(\vu\right)
    }{
      \partial \vlambda
    }
    \nabla f\left( \vt_{\vlambda}\left(\vu\right) \right)
    \nonumber
    \\
    &\;=
    {\vg_{f}^{\top}}
    {\left(
      \frac{
        \partial \vt_{\vlambda}\left(\vu\right)
      }{
        \partial \vlambda
      }
    \right)}^{\top}
    \frac{
      \partial \vt_{\vlambda}\left(\vu\right)
    }{
      \partial \vlambda
    }
    \vg_{f}.\label{eq:variational_gradient_norm_identity_eq1}
  \end{alignat}
  Naturally, the derivative of the reparameterization function will depend on the specific parameterization used.

  \paragraph{Proof for Cholesky}

  Let \(p\) denote the number of scalar variational parameters such that \(\vlambda = (\lambda_1, \ldots, \lambda_p)\).
  Then,
  \begin{alignat*}{2}
    &{\left(
      \frac{
        \partial \vt_{\vlambda}\left(\vu\right)
      }{
        \partial \vlambda
      }
    \right)}^{\top}
    \frac{
      \partial \vt_{\vlambda}\left(\vu\right)
    }{
      \partial \vlambda
    }
    \\
    &\;=
    \sum^{d}_{i=1} 
    \frac{
      \partial \vt_{\vlambda}\left(\vu\right)
    }{
      \partial m_i
    }
    {\left(
    \frac{
      \partial \vt_{\vlambda}\left(\vu\right)
    }{
      \partial m_i
    }
    \right)}^{\top}
    +
    \sum^{d}_{i=1} 
    \sum^{d}_{j \leq i} 
    \frac{
      \partial \vt_{\vlambda}\left(\vu\right)
    }{
      \partial \lambda_{C_{ij}}
    }
    {\left(
    \frac{
      \partial \vt_{\vlambda}\left(\vu\right)
    }{
      \partial \lambda_{C_{ij}}
    }
    \right)}^{\top},
  \end{alignat*}
  where \(\lambda_{C_{ij}}\) denote the parameter responsible for the \(ij\)-th entry of \(\mC\), \(C_{ij}\).
  Notice that, unlike for the matrix square root parameterization~\citep{domke_provable_2019}, the sum for \(C_{ij}\) is only over the lower triangular section.

  For the derivatives with respect to \(m_i\) and \(C_{ij}\), \citet{domke_provable_2020, domke_provable_2019} show that
  \begin{alignat}{2}
    \frac{\partial \vt_{\vlambda}\left(\vu\right) }{ \partial m_i }   &= \boldupright{e}_i \quad
    \frac{\partial \vt_{\vlambda}\left(\vu\right) }{ \partial C_{ij} } &= \boldupright{e}_i u_j,\label{eq:covariance_derivative}
  \end{alignat}
  where \(\boldupright{e}_i\) is the unit basis of the \(i\)th component.

  Therefore,
  \begin{alignat}{2}
    &{\left(
      \frac{
        \partial \vt_{\vlambda}\left(\vu\right)
      }{
        \partial \vlambda
      }
    \right)}^{\top}
    \frac{
      \partial \vt_{\vlambda}\left(\vu\right)
    }{
      \partial \vlambda
    }
    \nonumber
    \\
    &\;=
    \sum^{d}_{i=1} 
    \boldupright{e}_i
    \boldupright{e}_i^{\top}
    +
    \sum^{d}_{i=1} 
    \sum_{j \leq i} 
    \frac{
      \partial \vt_{\vlambda}\left(\vu\right)
    }{
      \partial \lambda_{C_{ij}}
    }
    {\left(
    \frac{
      \partial \vt_{\vlambda}\left(\vu\right)
    }{
      \partial \lambda_{C_{ij}}
    }
    \right)}^{\top}
    \nonumber
    \\
    &\;=
    \boldupright{I}
    +
    \underbrace{
    \sum^{d}_{i=1} 
    \frac{
      \partial \vt_{\vlambda}\left(\vu\right)
    }{
      \partial \lambda_{C_{ii}}
    }
    {\left(
    \frac{
      \partial \vt_{\vlambda}\left(\vu\right)
    }{
      \partial \lambda_{C_{ii}}
    }
    \right)}^{\top}
    }_{\text{diagonal of \(\mC\)}}
    \nonumber
    \\
    &\qquad+
    \underbrace{
    \sum^{d}_{i=1} 
    \sum_{j < i} 
    \frac{
      \partial \vt_{\vlambda}\left(\vu\right)
    }{
      \partial \lambda_{C_{ij}}
    }
    {\left(
    \frac{
      \partial \vt_{\vlambda}\left(\vu\right)
    }{
      \partial \lambda_{C_{ij}}
    }
    \right)}^{\top}}_{\text{off-diagonal of \(\mC\)}},
    \label{eq:thm:variational_gradient_norm_identity_eq2}
  \end{alignat}
  leaving us with the derivatives of the scale term.

  The gradient with respect to \(\lambda_{C_{ij}}\), however, depends on the parameterization.
  That is,
  \begin{alignat}{2}
    \frac{
      \partial \vt_{\vlambda}\left(\vu\right)
    }{
      \partial \lambda_{C_{ij}}
    }
    &=
    \frac{
      \partial \vt_{\vlambda}\left(\vu\right)
    }{
      \partial C_{ij}
    }
    \frac{
      \partial C_{ij}
    }{
      \partial \lambda_{C_{ij}}
    }
    &=
    \boldupright{e}_i u_j
    \frac{
      \partial C_{ij}
    }{
      \partial \lambda_{C_{ij}}
    }. \label{eq:thm:variational_gradient_norm_identity_covderivative}
  \end{alignat}

  For the diagonal elements, \(\lambda_{C_{ii}} = s_i\).
  Thus,
  \begin{align}
    \frac{
      \partial C_{ii}
    }{
      \partial s_i
    }
    =
    \frac{
      \partial \phi\left(s_i\right)
    }{
      \partial s_i
    }
    =
    \phi^{\prime}\left(s_i\right). \label{eq:thm:variational_gradient_norm_identity_diag}
  \end{align}
  And for the off-diagonal elements, \(\lambda_{L_{ij}} = L_{ij}\), and
  \begin{align}
    \frac{
      \partial C_{ij}
    }{
      \partial L_{ij}
    }
    =
    1. \label{eq:thm:variational_gradient_norm_identity_offdiag}
  \end{align}

  Plugging \cref{eq:thm:variational_gradient_norm_identity_diag,eq:thm:variational_gradient_norm_identity_offdiag,eq:thm:variational_gradient_norm_identity_covderivative} into \cref{eq:thm:variational_gradient_norm_identity_eq2},
  \begin{alignat}{2}
    &{\left(
      \frac{
        \partial \vt_{\vlambda}\left(\vu\right)
      }{
        \partial \vlambda
      }
    \right)}^{\top}
    \frac{
      \partial \vt_{\vlambda}\left(\vu\right)
    }{
      \partial \vlambda
    }
    \nonumber
    \\
    &\;=
    \boldupright{I}
    +
    \underbrace{
    \sum^{d}_{i=1} 
    {\left( u_i \phi^{\prime}\left(s_i\right) \right)}^2
    \boldupright{e}_i \boldupright{e}_i^{\top}
    }_{\text{diagonal of \(\mC\)}}
    +
    \underbrace{
    \sum^{d}_{i=1} 
    \sum_{j=1, j < i} 
    u_j^2 \, \boldupright{e}_i \boldupright{e}_i^{\top}
    }_{\text{off-diagonal of \(\mC\)}}
    \nonumber
    \\
    &\;=
    \boldupright{I}
    +
    \underbrace{
    \sum^{d}_{i=1} 
    u_i^2 {\left(\phi^{\prime}\left(s_i\right) \right)}^2
    \boldupright{e}_i \boldupright{e}_i^{\top}
    }_{\text{diagonal of \(\mC\)}}
    +
    \underbrace{
    \sum^{d}_{i=1} 
    \sum_{j \leq i} 
    u_j^2 \, \boldupright{e}_i \boldupright{e}_i^{\top}
    -
    \sum^{d}_{i=1} 
    u_i^2 \, \boldupright{e}_i \boldupright{e}_i^{\top}
    }_{\text{off-diagonal of \(\mC\)}}
    \nonumber
    \\
    &\;=
    \boldupright{I}
    +
    \underbrace{
      \mU \, \mPhi
    }_{\text{diagonal of \(\mC\)}}
    +
    \underbrace{
    \mSigma
    -
    \mU
    }_{\text{off-diagonal of \(\mC\)}}
    \nonumber
    \\
    &\;=
    \left( \boldupright{I} + \mSigma \right)
    +
    \mU \left( \mPhi - \boldupright{I} \right), \label{eq:variational_gradient_norm_identity_jacinner}
  \end{alignat}
  where \(\mU,\mPhi,\mSigma\) are diagonal matrices defined as
  \begin{alignat*}{2}
    &
    \mPhi
    &&=
    \mathrm{diag}\left(
    \left[ {\phi^{\prime}\left(s_1\right)}^2, \ldots , {\phi^{\prime}\left(s_d\right)}^2 \right]
    \right)
    \\
    &\mU
    &&=
    \mathrm{diag}\left(
    \left[ u_1^2, \ldots, u_d^2 \right]
    \right)
    \\
    &\mSigma
    &&=
    \mathrm{diag}\left(
    \left[ u_1^2, u_1^2 + u_2^2,\, \ldots\,, {\textstyle\sum^{d}_{i=1} u_i^2} \right]
    \right).
  \end{alignat*}
  The major difference with the proof of \citet[Lemma 8]{domke_provable_2019} for the matrix square root case is that we only sum the \(u_j^2 \boldupright{e}_i \boldupright{e}_i^{\top}\) terms over the \textit{lower diagonal elements}.
  This is the variance reduction effect we get from using the Cholesky parameterization.

  Coming back to \cref{eq:variational_gradient_norm_identity_eq1}, 
  \begin{alignat}{2}
    &\norm{\nabla_{\vlambda} f\left( \vt_{\vlambda}\left(\vu\right) \right)}_2^2
    \nonumber
    \\
    \;&=
      \vg_f^{\top}
      {\left(
        \frac{
          \partial \vt_{\vlambda}\left(\vu\right)
        }{
          \partial \vlambda
        }
        \right)}^{\top}
      \frac{
        \partial \vt_{\vlambda}\left(\vu\right)
      }{
        \partial \vlambda
      }
      \vg
    \nonumber
    \\
    &=
      \vg_{f}^{\top}
      \Big(
        \left( \boldupright{I} + \mSigma \right)
        +
        \mU \left( \mPhi - \boldupright{I} \right)
      \Big)
      \vg_{f}
    \nonumber
    \\
    &=
    {\lVert \vg_{f} \rVert}_2^2
    +
    \vg_{f}^{\top} \mSigma \vg_{f}
    +
    \vg_{f}^{\top}
    \mU \left( \mPhi - \boldupright{I} \right)
    \vg_{f}.
    \label{eq:variational_gradient_norm_identity_conclusion}
  \end{alignat}

  \paragraph{Proof for Mean-field}
  For the mean-field variational family, the covariance has only diagonal elements.
  Therefore, \cref{eq:variational_gradient_norm_identity_jacinner} becomes
  \begin{alignat}{2}
    {\left(
      \frac{
        \partial \vt_{\vlambda}\left(\vu\right)
      }{
        \partial \vlambda
      }
    \right)}^{\top}
    \frac{
      \partial \vt_{\vlambda}\left(\vu\right)
    }{
      \partial \vlambda
    }
    &=
    \boldupright{I} + \mU \mPhi,
    \nonumber
  \end{alignat}
  and \cref{eq:variational_gradient_norm_identity_conclusion} becomes
  \begin{alignat}{2}
    \norm{\nabla_{\vlambda} f\left( \vt_{\vlambda}\left(\vu\right) \right)}_2^2
    =
      \vg_{f}^{\top}
      \left(
        \boldupright{I}
        +
        \mU \mPhi
      \right)
      \vg_{f}
    =
    {\lVert \vg_{f} \rVert}_2^2
    +
    \vg_{f}^{\top}
    \mU \mPhi
    \vg_{f}.
    \nonumber
  \end{alignat}
\end{proofEnd}

Note that the relationships in this lemma are all equalities, which can be bounded with known quantities, as done in the next lemma.
We note here that if any of our analyses were to be improved, this shall by done by obtaining tighter bounds on the equalities in \cref{thm:general_variational_gradient_norm_identity}.

\begin{theoremEnd}[\keylemmaproofoption,category=upperboundkeylemmagradientnormbound]{lemma}\label{thm:general_variational_gradient_norm_bound}
Let \(\vt_{\vlambda}: \mathbb{R}^d \rightarrow \mathbb{R}^d\) be a location-scale reparameterization function (\cref{def:reparam}), \(f : \mathbb{R}^d \rightarrow \mathbb{R} \) be a differentiable function, and let \(\phi\) satisfy \cref{assumption:phi_lipschitz}.
  \vspace{-5ex}
  \begin{enumerate}[label=(\roman*)]
    \setlength\itemsep{-1ex}
    \item Mean-Field
    {%
    \setlength{\belowdisplayskip}{1ex} \setlength{\belowdisplayshortskip}{1ex}%
    \setlength{\abovedisplayskip}{1ex} \setlength{\abovedisplayshortskip}{1ex}%
      \[
        \norm{\nabla_{\vlambda} f\left( \vt_{\vlambda}\left(\vu\right) \right)}_2^2
        \leq
        \left(1 + \norm{ \mU }_{\mathrm{F}} \right)
        {\lVert \nabla f\left( \vt_{\vlambda}\left(\vu\right) \right) \rVert}_2^2,
      \]
      where \(\mU\) is a diagonal matrix such that \(U_{ii} = u_i^2\).
    }%
    \item Cholesky
    {%
    \setlength{\belowdisplayskip}{1ex} \setlength{\belowdisplayshortskip}{1ex}%
    \setlength{\abovedisplayskip}{1ex} \setlength{\abovedisplayshortskip}{1ex}%
      \[
        \norm{\nabla_{\vlambda} f\left( \vt_{\vlambda}\left(\vu\right) \right)}_2^2
        \leq
        \left(1 + \norm{ \vu }_2^2 \right)
        {\lVert \nabla f\left( \vt_{\vlambda}\left(\vu\right) \right) \rVert}_2^2,
      \]
      }%
      where the equality holds for the matrix square root parameterization.
  \end{enumerate}
\end{theoremEnd}
\vspace{-2ex}
\begin{proofEnd}
  The proof continues from the result of \cref{thm:general_variational_gradient_norm_identity}.

  \paragraph{Proof for Cholesky}
  \cref{thm:general_variational_gradient_norm_identity} shows that
  \begin{alignat*}{2}
    \norm{ \nabla_{\vlambda} f\left( \vt_{\vlambda}\left(\vu\right) \right) }_2^2
    = 
    {\lVert \vg_{f} \rVert}_2^2
    +
    \vg_{f}^{\top} \mSigma \vg_{f}
    +
    \vg_{f}^{\top}
    \mU \left( \mPhi - \boldupright{I} \right)
    \vg_{f},
  \end{alignat*}
  where \(\vg_f = \nabla f\left(\vt_{\vlambda}\left(\vu\right)\right) \).

  By the 1-Lipschitz assumption, the entries of the diagonal matrix \(\Phi\) satisfy
  \begin{align*}
    \Phi_{ii} = {\phi^{\prime}\left(d_i\right)}^2 \leq 1,
  \end{align*}
  which means
  \begin{alignat*}{2}
    \mPhi \preceq \boldupright{I}
    \;\Rightarrow\;
    \mU \left( \mPhi - \boldupright{I} \right)
    \preceq
    0
    \;\Rightarrow\;
    {\vg_{f}}^{\top} \mU \left( \mPhi - \boldupright{I} \right) \vg_{f} \leq 0.
  \end{alignat*}
  Therefore, for the full-rank Cholesky parameterization and a 1-Lipschitz conditioner \(\phi\),
  \begin{alignat*}{2}
    &\norm{\nabla_{\vlambda} f\left( \vt_{\vlambda}\left(\vu\right) \right)}_2^2
    \\
    &\;=
    {\lVert \nabla f\left( \vt_{\vlambda}\left(\vu\right) \right) \rVert}_2^2
    +
    {\vg_{f}}^{\top}
    \mSigma
    \vg_{f}
    +
    {\vg_{f}}^{\top} \mU \left( \mPhi - \boldupright{I} \right) \vg_{f}
    \\
    &\;\leq
    {\lVert \nabla f\left( \vt_{\vlambda}\left(\vu\right) \right) \rVert}_2^2
    +
    {\vg_{f}}^{\top}
    \mSigma
    \vg_{f}
    \\
    &\;\leq
    {\lVert \nabla f\left( \vt_{\vlambda}\left(\vu\right) \right) \rVert}_2^2
    +
    \norm{ \mSigma }_{2,2} {\lVert \nabla f\left( \vt_{\vlambda}\left(\vu\right) \right) \rVert}_2^2
    \\
    &\;=
    {\lVert \nabla f\left( \vt_{\vlambda}\left(\vu\right) \right) \rVert}_2^2
    +
    \left( \sum^{d}_{i=1} u_{i}^2 \right)  {\lVert \nabla f\left( \vt_{\vlambda}\left(\vu\right) \right) \rVert}_2^2
    \\
    &\;=
    \left(1 + \norm{\vu}^2_2\right) {\lVert \nabla f\left( \vt_{\vlambda}\left(\vu\right) \right) \rVert}_2^2,
  \end{alignat*}
  where \(\norm{\mU}_{2,2}\) is the \(L_2\) operator norm of \(\mU\).
  This upper bound coincides with that of the matrix square root parameteration.
  Thus, unforunately, this bound fails to acknowledge the lower variance of the Cholesky parameterization, coinciding with that of the matrix square root parameterization.

  \paragraph{Proof for Mean-field (\cref{def:meanfield})}
  For the mean-field parameterization,~\cref{thm:general_variational_gradient_norm_identity} shows that
  \begin{alignat*}{2}
    \norm{ \nabla_{\vlambda} f\left( \vt_{\vlambda}\left(\vu\right) \right) }_2^2
    =
    {\lVert \vg_{f} \rVert}_2^2
    +
    \vg_{f}^{\top}
    \mU \mPhi
    \vg_{f}.
  \end{alignat*}

  For the second term,  
  \begin{alignat*}{2}
    \vg_{f}^{\top}
    \mU \mPhi
    \vg_{f}
    \leq
    {\lVert \mU \rVert}_{2,2} {\lVert \mPhi \rVert}_{2,2}
    {\lVert \vg_{f} \rVert}^2_2.
  \end{alignat*}
  By the \(1\)-Lipschitzness of \(\phi\),
  \[
    {\lVert \mPhi \rVert}_{2,2}
    = \sigma_{\mathrm{max}}\left( \mPhi \right)
    = \max_{i = 1, \ldots, d} {\phi^{\prime}\left( s_i \right)}^2
    \leq 1.
  \]
  Then,
  \begin{alignat}{2}
    \vg_{f}^{\top}
    \left( \mU \mPhi \right)
    \vg_{f}
    &\leq
    {\lVert \mU \rVert}_{2,2} \,
    {\lVert \vg_{f} \rVert}^2_2 \label{eq:variational_gradient_norm_identity_mf_eq1}
    \\
    &\leq
    {\lVert \mU \rVert}_{\mathrm{F}} \,
    {\lVert \vg_{f} \rVert}^2_2, \label{eq:variational_gradient_norm_identity_mf_eq2}
  \end{alignat}
  which gives the result.
  Here, unlike the bounds on \(\mPhi\), the bounds in \cref{eq:variational_gradient_norm_identity_mf_eq1,eq:variational_gradient_norm_identity_mf_eq2} are quite loose, and become looser as the dimensionality increases.

\end{proofEnd}


\cref{thm:general_variational_gradient_norm_identity} act as the interface between the properties of the parameterization and the likelihood \(f\).

\begin{remark}[\textbf{Variance Reduction Through \(\phi\)}]
  A \textit{nonlinear} Cholesky parameterization with a 1-Lipschitz \(\phi\) achieves lower or equal variance compared to the matrix square root and \textit{linear} Cholesky, where the equality is achieved with the matrix square root parameterization.
\end{remark}

\vspace{-2.0ex}%
\paragraph{Dimension Dependence of Mean-Field}
The superior dimensional dependence of the mean-field parameterization is given by the following lemma:

\begin{theoremEnd}[\keylemmaproofoption,category=upperboundkeylemmameanfield]{lemma}\label{thm:meanfield_u_identity}
  Let the assumptions of~\cref{thm:general_variational_gradient_norm_bound} hold and \(\rvvu \sim \varphi\) satisfy \cref{assumption:symmetric_standard}.
  Then, for the mean-field parameterization,
    {
    \setlength{\belowdisplayskip}{1.ex} \setlength{\belowdisplayshortskip}{1.ex}
    \setlength{\abovedisplayskip}{1.ex} \setlength{\abovedisplayshortskip}{1.ex}
  \begin{alignat*}{2}
    &\mathbb{E}\norm{\vt_{\vlambda}\left(\rvvu\right) - \vz}_2^2 \left(1 + \norm{\mathbfsfit{U}}_{\mathrm{F}} \right)
    \\
    &\quad\leq
    \left(\sqrt{d \kappa} + \kappa\sqrt{d} + 1\right) \, \norm{ \vm - \vz }_2^2
    +
    \left(2 \kappa \sqrt{d} + 1\right)
    \norm{\mC}_{\mathrm{F}}^2.
  \end{alignat*}
  }
\end{theoremEnd}
\vspace{-1ex}
\begin{proofEnd}
  The key idea is to prove a similar result as \cref{thm:reparam_u_identity}, but with better constants to reflect that the mean-field parameterization has a lower variance. 

  First,
  \begin{align}
    &\mathbb{E}\norm{\vt_{\vlambda}\left(\rvvu\right) - \vz}_2^2 \, \left( 1 + \norm{\mathbfsfit{U}}_{\mathrm{F}} \right) \nonumber \\
    &\;=
    \mathbb{E}\norm{\vt_{\vlambda}\left(\rvvu\right) - \vz}_2^2 
    + \mathbb{E} \norm{\mathbfsfit{U}}_{\mathrm{F}} \, \norm{\vt_{\vlambda}\left(\rvvu\right) - \vz}_2^2, 
    \nonumber
\shortintertext{applying \cref{thm:reparam_quadratic},}
    &\;=
    \norm{ \vm - \vz }_2^2 + \norm{\mC}_{\mathrm{F}}
    +
    \mathbb{E} \norm{\mathbfsfit{U}}_{\mathrm{F}} \, \norm{\vt_{\vlambda}\left(\rvvu\right) - \vz}_2^2.
    \label{thm:meanfield_eq0}
  \end{align}
  
  The last term decomposes as 
  \begin{alignat}{2}
    \mathbb{E} \norm{\mathbfsfit{U}}_{\mathrm{F}} \norm{\vt_{\vlambda}\left(\rvvu\right) - \vz}_2^2
    \nonumber
    &=
    \underbrace{
      \mathbb{E} \norm{\mathbfsfit{U}}_{\mathrm{F}} \, \rvvu^{\top} \mC^{\top} \mC \rvvu 
    }_{\text{Term \ding{182}}}
    \nonumber
    \\
    &\quad+
    2\, 
    \underbrace{
      \mathbb{E} \norm{\mathbfsfit{U}}_{\mathrm{F}} \, \rvvu^{\top} \mC^{\top} \left( \vm - \vz \right)
    }_{\text{Term \ding{183}}}
    \nonumber
    \\
    &\quad+
    \underbrace{\mathbb{E} \norm{\mathbfsfit{U}}_{\mathrm{F}}}_{\text{Term \ding{184}}} \, \norm{ \vm - \vz }_2^2.
    \label{thm:meanfield_eq1}
  \end{alignat}
  We will now focus on the stochastic terms \ding{182}-\ding{184} one by one.
  
  First, for Term \ding{182}, notice that the mean-field parameterization implies that \(\mC = \mathrm{diag}\left( c_1, \ldots, c_d \right)\).
  Thus, 
  \begin{alignat}{2}
    \mathbb{E} \norm{\mathbfsfit{U}}_{\mathrm{F}} \,  \rvvu^{\top} \mC^{\top} \mC \rvvu 
    &=
    \mathbb{E} \left( \sqrt{ \sum_{i=1}^d \rvu_i^4 } \right) \left( \sum_{i=1}^d c_i^2 \, u_i^2 \right)
    \nonumber
    \\
    &=
    \sum_{i=1}^d c_i^2 \, \mathbb{E} \left( \sqrt{ \sum_{j=1}^d \rvu_j^4 } \right) \rvu_i^2,
    \nonumber
\shortintertext{applying Cauchy-Schwarz inequality for expectations,}
    &\leq
    \sum_{i=1}^d c_i^2 \sqrt{ \left( \mathbb{E} \sum_{j=1}^d \rvu_j^4  \right) \left( \mathbb{E}  \rvu_i^4 \right) }
    \nonumber
\shortintertext{and given \cref{assumption:symmetric_standard},}
    &\;=
    \sum_{i=1}^d c_i^2 \, \sqrt{ d \kappa^2 }
    \nonumber
    \\
    &\;=
    \kappa \sqrt{d} \, \norm{ \mC }_{\mathrm{F}}^2.
    \label{thm:meanfield_eq5}
  \end{alignat}
  
  Term \ding{183} can be bounded as 
  \begin{alignat}{2}
    &\mathbb{E} 
      \norm{\mathbfsfit{U}}_{\mathrm{F}} \, \rvvu^{\top} 
    \mC^{\top} \left( \vm - \vz \right) 
    \nonumber
\shortintertext{using the Cauchy-Schwarz inequality for vectors as}
    &\;\leq
    \mathbb{E} \norm{\mathbfsfit{U}}_{\mathrm{F}} \, \norm{\mC \rvvu}_2 \norm{ \vm - \vz }_2,
    \nonumber
\shortintertext{again, applying the inequality for expectations,}
    &\;=
    \sqrt{
    \mathbb{E}
    \norm{\mathbfsfit{U}}_{\mathrm{F}}^2 \,
    \mathbb{E}
    \norm{\mC \rvvu}_2^2 
    } \,
    \norm{ \vm - \vz }_2
    \nonumber
    \\
    &\;=
    \sqrt{
    \mathbb{E}
    \left(
    \sum^d_{i=1} \rvu_i^4  
    \right) \,
    \mathrm{tr}\left(\mC^{\top} \mC \, \mathbb{E} \rvvu \rvvu^{\top} \right)
    } \,
    \norm{ \vm - \vz }_2,
    \nonumber
\shortintertext{from \cref{assumption:symmetric_standard},}
    &\;=
    \sqrt{
    d \kappa \,
    \mathrm{tr}\left(\mC^{\top} \mC \right)
    } \,
    \norm{ \vm - \vz }_2
    \nonumber
    \\
    &\;=
    \sqrt{d \kappa} \,
    \norm{\mC}_{\mathrm{F}} \,
    \norm{ \vm - \vz }_2
     \nonumber
    \\
    &\;=
    \sqrt{d \kappa} \,
    \sqrt{ \norm{\mC}_{\mathrm{F}}^2 \,
           \norm{ \vm - \vz }_2^2 },
    \nonumber
\shortintertext{and by the arithmetic mean-geometric mean inequality,}
    &\;=
    \frac{\sqrt{d \kappa}}{2}
    \left(
      \norm{\mC}_{\mathrm{F}}^2
      +
      \norm{ \vm - \vz }_2^2
    \right).
    \label{thm:meanfield_eq3}
  \end{alignat}

  Finally, Term \ding{184} follows as
  \begin{alignat}{2}
    \mathbb{E} \norm{\mathbfsfit{U}}_{\mathrm{F}}
    &=
    \mathbb{E} \sqrt{ \sum_{i=1}^d \rvu_i^4  },
    \nonumber
\shortintertext{using Jensen's inequality,}
    &\leq
    \sqrt{ \mathbb{E} \sum_{i=1}^d \rvu_i^4 }
    \nonumber
    \\
    &=
    \sqrt{ d \kappa }.
    \label{thm:meanfield_eq2}
  \end{alignat}

  Combining all the results, \cref{thm:meanfield_eq0} becomes
  \begin{alignat*}{2}
    &\mathbb{E}\norm{\vt_{\vlambda}\left(\rvvu\right) - \vz}_2^2 \, \left( 1 + \norm{\mathbfsfit{U}}_{\mathrm{F}} \right)
    \\
    &\;\leq
    \norm{ \vm - \vz }_2^2 + \norm{\mC}_{\mathrm{F}}^2
    \\ 
    &\;\quad+
    \mathbb{E} \norm{\mathbfsfit{U}}_{\mathrm{F}} \, \rvvu^{\top} \mC^{\top} \mC \rvvu
    \\ 
    &\;\quad+
    2\,\mathbb{E} \norm{\mathbfsfit{U}}_{\mathrm{F}} \, \rvvu^{\top} \mC^{\top} \norm{ \vm - \vz }_2^2
    \\ 
    &\;\quad+
    \mathbb{E} \norm{\mathbfsfit{U}}_{\mathrm{F}} \, \norm{ \vm - \vz }_2^2
\shortintertext{and applying \cref{thm:meanfield_eq2,thm:meanfield_eq3,thm:meanfield_eq5},}
    &\;\leq
    \norm{ \vm - \vz }_2^2 + \norm{\mC}_{\mathrm{F}}^2
    \\ 
    &\;\quad+
    \kappa\sqrt{d} \norm{\mC}_{\mathrm{F}}
    \\ 
    &\;\quad+
    \kappa \sqrt{d} \left(
     \norm{\mC}_{\mathrm{F}}^2 + \norm{\vm - \vz}_2^2
    \right)
    \\ 
    &\;\quad+
    \sqrt{d \kappa} \, \norm{ \vm - \vz }_2^2
    \\ 
    &\;=
    \left(\sqrt{d \kappa} + \kappa\sqrt{d} + 1\right) \, \norm{ \vm - \vz }_2^2
    +
    \left(2 \kappa \sqrt{d} + 1\right)
    \norm{\mC}_{\mathrm{F}}^2.
  \end{alignat*}

\end{proofEnd}


\begin{remark}[\textbf{Superior Variance of Mean-Field}]\label{remark:meanfield_superiority}
  The mean-field parameterization has {\small\(\mathcal{O}\left(\sqrt{d}\right)\)} dimensional dependence compared to the \(\mathcal{O}\left(d\right)\) dimensional dependence of the full-rank parameterizations in \cref{thm:reparam_u_identity}.
\end{remark}

Lastly, the following lemma is the basic building block for all of our upper bounds:

\begin{theoremEnd}[\keylemmaproofoption,category=upperboundkeylemmavariancegeneral]{lemma}\label{thm:gradient_variance_general_upper_bound}
  Let \(\vg_{M}\) be the \(M\)-sample gradient estimator of \(F\) (\cref{def:generic_elbo}) for some function \(f,h\) and let \(\rvvu\) be some random variable.
  Then, 
  {%
  \setlength{\belowdisplayskip}{1ex} \setlength{\belowdisplayshortskip}{1ex}%
  \setlength{\abovedisplayskip}{1ex} \setlength{\abovedisplayshortskip}{1ex}%
  \begin{align*}
    \mathbb{E} \norm{\rvvg_M }^2_2
    \leq
    \frac{1}{M} \mathbb{E}{ \norm{
      \nabla_{\vlambda} f\left(\vt_{\vlambda}\left(\rvvu\right)\right)
      }_2^2
    }
    + \norm{ \nabla F\left(\vlambda\right) }^2_2.
  \end{align*}
  }%
\end{theoremEnd}
\vspace{-1ex}
\begin{proofEnd}
  From the definition of variance,
  \begin{alignat}{2}
    &\mathbb{E} \norm{\vg_M }^2_2
    \nonumber
    \\
    &\;=
    \mathrm{tr}\,\V{ \vg_M } + \norm{\mathbb{E} \vg_M }^2_2,
    \nonumber
\shortintertext{following the definition in \cref{eq:def_gradient_M_est},}
    &\;=
    \mathrm{tr}\,\V{ \frac{1}{M} \sum_{m=1}^M \vg_m } + \norm{ \nabla F\left(\vlambda\right) }^2_2,
    \nonumber
\shortintertext{and then the definition in \cref{eq:def_gradient_m_est},}
    &\;=
    \mathrm{tr}\,\V{
      \frac{1}{M} \sum_{m=1}^M \nabla_{\vlambda} f\left(\vt_{\vlambda}\left(\rvvu_m\right)\right) + \nabla h\left(\vlambda\right)
    }
    + \norm{ \nabla F\left(\vlambda\right) }^2_2,
    \nonumber
\shortintertext{by the linearity of variance,}
    &\;=
    \frac{1}{M} \mathrm{tr}\,\V{
      \nabla_{\vlambda} f\left(\vt_{\vlambda}\left(\rvvu\right)\right)
    }
    + \norm{ \nabla F\left(\vlambda\right) }^2_2
    \nonumber
    \\
    &\;=
    \frac{1}{M} \left(
    \mathbb{E}{ \norm{\nabla_{\vlambda} f\left(\vt_{\vlambda}\left(\rvvu\right)\right)}_2^2 }
    -
    \norm{ \mathbb{E}{ \nabla_{\vlambda} f\left(\vt_{\vlambda}\left(\rvvu\right)\right)} }_2^2
    \right)
    \nonumber
    \\
    &\qquad+ \norm{ \nabla F\left(\vlambda\right) }^2_2
    \label{eq:thm_gradient_variance_general_definition}
    \\
    &\;\leq
    \frac{1}{M} \mathbb{E}{ \norm{
      \nabla_{\vlambda} f\left(\vt_{\vlambda}\left(\rvvu\right)\right)
      }_2^2
    }
    + \norm{ \nabla F\left(\vlambda\right) }^2_2.
    \nonumber
  \end{alignat}
\end{proofEnd}


\begin{theoremEnd}[\lemmaproofoption, category=upperboundlemma]{lemma}\label{thm:reparam_quadratic}
  Let \(\vt_{\vlambda}: \mathbb{R}^d \rightarrow \mathbb{R}^d\) be a location-scale reparameterizaiton function (\cref{def:reparam}).
  Also, let \(\vz \in \mathbb{R}^d\) be some vector, and let \(\rvvu \sim \varphi\) satisfy~\cref{assumption:symmetric_standard}.
  Then,
  \begin{alignat*}{2}
    \mathbb{E}\norm{\vt_{\vlambda}\left(\rvvu\right) - \vz}_2^2
    &=
    \norm{\vm - \vz}^2_2 + \norm{\mC}_{\mathrm{F}}^2.
  \end{alignat*}
\end{theoremEnd}
\begin{proofEnd}
  \begin{alignat}{2}
    \mathbb{E}\norm{ \vt_{\vlambda}\left(\rvvu\right) - \vz }_2^2
    &=
    \mathbb{E}\norm{ \mC \rvvu + \vm - \vz }_2^2
    \nonumber
    \\
    &=
    \mathbb{E} \rvvu^{\top} \mC^{\top} \mC \rvvu + 2\,\mathbb{E} \rvvu^{\top} \mC^{\top} \vm - 2\,\mathbb{E} \rvvu^{\top} \mC^{\top} \vz
    \nonumber
    \\
    &\;\quad + \vm^{\top} \vm - 2\,\vm^{\top} \vz + \vz^{\top} \vz.
    \label{eq:reparam_quadratic_eq1}
  \end{alignat}
  The first three terms follow as
  \begin{alignat*}{2}
    &\mathbb{E} \rvvu^{\top} \mC^{\top} \mC \rvvu + 2\,\mathbb{E} \rvvu^{\top} \mC^{\top} \vm - 2\,\mathbb{E} \rvvu^{\top} \mC^{\top} \vz
    \\
    &\;=
    \mathbb{E} \mathrm{tr}\left(\rvvu^{\top} \mC^{\top} \mC \rvvu\right) + 2\,\mathbb{E} \rvvu^{\top} \mC^{\top} \vm -  2\,\mathbb{E} \rvvu^{\top} \mC^{\top} \vz
    \\
    &\;=
    \mathrm{tr}\left( \mC^{\top} \mC \mathbb{E} \rvvu \rvvu^{\top}\right) + 2\,\mathbb{E} \rvvu^{\top} \mC^{\top} \vm -  2\,\mathbb{E} \rvvu^{\top} \mC^{\top} \vz,
\shortintertext{applying \cref{thm:u_identities},}
    &\;=
    \mathrm{tr}\left( \mC^{\top} \mC \right)
    \\
    &\;=
    \norm{ \mC }_{\mathrm{F}}^2.
  \end{alignat*}
  Applying this to \cref{eq:reparam_quadratic_eq1},
  \begin{alignat*}{2}
    \mathbb{E}\norm{\vt_{\vlambda}\left(\rvvu\right) - \vz}_2^2
    &=
    \vm^{\top} \vm - 2\,\vm^{\top} \vz + \vz^{\top} \vz + \norm{\mC}_{\mathrm{F}}^2
    \\
    &=
    \norm{\vm - \vz}^2_2 + \norm{\mC}_{\mathrm{F}}^2.
  \end{alignat*}
\end{proofEnd}

\subsection{Upper Bounds}\label{section:upper_bound}
\vspace{-.5ex}

We restrict our analysis to the class of log-likelihoods that satisfy the following conditions: 
\vspace{.5ex}
\begin{definition}[\textbf{\(L\)-smoothness}]\label{def:L_smoothness}
  A function \(f : \mathbb{R}^d \rightarrow \mathbb{R}\) is \(L\)-smooth if it satisfies the following for all \(\vzeta, \vzeta^{\prime} \in \mathbb{R}^d\):
  {\small%
  \setlength{\belowdisplayskip}{1.ex} \setlength{\belowdisplayshortskip}{1.ex}%
  \setlength{\abovedisplayskip}{1.ex} \setlength{\abovedisplayshortskip}{1.ex}%
  \begin{align*}
    {\lVert \nabla f\left(\vzeta\right) - \nabla f\left(\vzeta^{\prime}\right) \rVert}_2
    \leq
    L \, {\lVert \vzeta - \vzeta^{\prime} \rVert}_2.
  \end{align*}
  }%
\end{definition}

\vspace{.5ex}
\begin{definition}[\textbf{Quadratic Functional Growth}]\label{def:quadratic_growth}
  A function \(f : \mathbb{R}^d \rightarrow \mathbb{R}\) is \(\mu\)-quadratically growing if
  {\small%
  \setlength{\belowdisplayskip}{1.ex} \setlength{\belowdisplayshortskip}{1.ex}%
  \setlength{\abovedisplayskip}{1.ex} \setlength{\abovedisplayshortskip}{1.ex}%
  \begin{align*}
   \frac{\mu}{2} {\lVert \vzeta - \bar{\vzeta} \rVert}_2^2
    \leq
    f\left(\vzeta\right) - f^*
  \end{align*}
  }%
  for all \(\vzeta \in \mathbb{R}^d\), where \(\bar{\vzeta} = \Pi_f\left(\vzeta\right) \) is a projection of \(\vzeta\) onto the set of minimizers of \(f\) and \(f^* = \inf_{\vzeta \in \mathbb{R}^d} f\left(\vzeta\right)\).
\end{definition}
\vspace{-1ex}
The quadratic growth condition has first been used by \citep{anitescu_degenerate_2000} and is strictly weaker than the Polyak-\L{}ojasiewicz inequality (see \citealt[Appendix A]{karimi_linear_2016} for the proof).
Furthermore, for \(\mu\)-strongly (quasar) convex functions~\citep{hinder_nearoptimal_2020,jin_convergence_2020} automatically satisfy quadratic growth, but our analysis does \textit{not} require (quasar) convexity.

Both assumptions are commonly used in SGD.
For studying the gradient variance of BBVI, assuming both smoothness and quadratic growth is weaker than the assumptions of \citet{xu_variance_2019} but stronger than those of~\citet{domke_provable_2019}, who assumed only smoothness.
The additional assumption on growth is necessary to extend his results to establish the \textit{ABC} condition.

For the variational family, we assume the followings:
\vspace{.5ex}
\begin{assumption}\label{assumption:q}
\(q_{\psi,\vlambda}\) is a member of the ADVI family (\cref{def:advi}), where the underlying  \(q_{\vlambda}\) is a member of the location-scale family (\cref{def:family}) with its base distribution \(\varphi\) satisfying \cref{assumption:symmetric_standard}.
\end{assumption}



\vspace{-1.ex}
\paragraph{Entropy-Regularized Form}
First, we provide the upper bound for the ELBO in entropy-regularized form.
This result does \textit{not} require any modifications to vanilla SGD.

\vspace{.5ex}

\begin{theoremEnd}[\theoremproofoption,category=upperboundtheorem]{theorem}\label{thm:gradient_upper_bound}
  Let \(\rvvg_{M}\) be an \(M\)-sample estimate of the gradient of the ELBO in entropy regularized form (\cref{def:entropy_form}).
  Also, assume that \cref{assumption:q,assumption:phi_lipschitz} hold,
  \begin{itemize}[leftmargin=3em]
    \vspace{-1.5ex}
    \setlength\itemsep{0ex}
    \item \(f_{\mathrm{H}}\) is \(L_{\mathrm{H}}\)-smooth, and
    \item \(f_{\mathrm{KL}}\) is \(\mu_{\mathrm{KL}}\)-quadratically growing.
    \vspace{-1.5ex}
  \end{itemize}
  Then, 
  {\small%
  \setlength{\belowdisplayskip}{1ex} \setlength{\belowdisplayshortskip}{1ex}%
  \setlength{\abovedisplayskip}{1ex} \setlength{\abovedisplayshortskip}{1ex}%
  \begin{align*}
    \hspace{-1.5em}
    \mathbb{E}\norm{\rvvg_{M}}_2^2
    &\leq
    \frac{4 L^2_{\mathrm{H}}}{\mu_{\mathrm{KL}} M} C\left(d, \kappa\right) \left( F\left(\vlambda\right) - F^* \right)
    + \norm{ \nabla F\left(\vlambda\right) }_2^2
    \\
    &\quad+ \frac{2 L^2_{\mathrm{H}}}{M} C\left(d, \kappa\right) {\lVert \bar{\vzeta}_{\mathrm{KL}} - \bar{\vzeta}_{\mathrm{H}} \rVert}_2^2
    \\
    &\quad+ \frac{4 L^2_{\mathrm{H}}}{\mu_{\mathrm{KL}} M} C\left(d, \kappa\right) \left( F^* - f_{\mathrm{KL}}^* \right),
  \end{align*}
  }%
  where
  {\small%
  \setlength{\belowdisplayskip}{1ex} \setlength{\belowdisplayshortskip}{1ex}%
  \setlength{\abovedisplayskip}{1ex} \setlength{\abovedisplayshortskip}{1ex}%
  \begin{alignat*}{2}
    C\left(d, \kappa\right) &= 2 \kappa \sqrt{d} + 1 &&\;\text{for mean-field,} \\
    C\left(d, \kappa\right) &= d + \kappa          &&\;\text{for the Cholesky and matrix square root,}
  \end{alignat*}
  }
  \(\bar{\zeta}_{\mathrm{KL}}\), \(\bar{\zeta}_{\mathrm{H}}\) are the stationary points of \(f_{\mathrm{KL}}\), \(f_{\mathrm{H}}\), respectively,
  \(F^* = \inf_{\vlambda \in \mathbb{R}^p} F\left(\vlambda\right)\), and \(f_{\mathrm{KL}}^* = \inf_{\vzeta \in \mathbb{R}^d} f\left(\zeta\right)\).
\end{theoremEnd}
\vspace{-1ex}
\begin{proofsketch}
  From \cref{thm:gradient_variance_general_upper_bound}, we can see that the key quantity of upper bounding the gradient variance is to analyze \(\mathbb{E} \norm{ \nabla_{\vlambda} f_{\mathrm{H}} \left( \vt_{\vlambda}\left(\rvvu\right) \right) } \).
  The bird's eye view of the proof is as follows:
  \begin{enumerate}
  \vspace{-1ex}
    \setlength\itemsep{-.5ex}
    \item[\ding{182}] The relationship between \( \norm{ \nabla_{\vlambda} f_{\mathrm{H}} \left( \vt_{\vlambda}\left(\rvvu\right) \right) }_2^2 \) and \( \norm{ \nabla f_{\mathrm{H}} \left( \vt_{\vlambda}\left(\rvvu\right) \right) }_2^2 \) is established through \cref{thm:general_variational_gradient_norm_bound}.
    \item[\ding{183}] Then, the \(L_{\mathrm{H}}\)-smoothness of \(f_{\mathrm{H}}\) relates \( \norm{ \nabla f_{\mathrm{H}} \left( \vt_{\vlambda}\left(\rvvu\right) \right) }_2^2 \) with \( {\lVert \vt_{\vlambda}\left(\rvvu\right) - \bar{\vzeta}_{\mathrm{H}} \rVert}_2^2\), the average squared distance from \(f_{\mathrm{H}}\)'s stationary point.
    \item[\ding{184}] The average squared distance enables the simplification of stochastic terms through \cref{thm:meanfield_u_identity,thm:reparam_u_identity}. This step also introduces dimension dependence.
  \vspace{-1ex}
  \end{enumerate}
  From here, we are now left with the \(\mathbb{E} {\lVert \vt_{\vlambda}\left(\rvvu\right) - \bar{\vzeta}_{\mathrm{H}} \rVert}_2^2\) term.
  One might be tempted to assume the quadratic growth assumption on \(f_{\mathrm{H}}\) and proceed as
  {%
  \setlength{\belowdisplayskip}{1.ex} \setlength{\belowdisplayshortskip}{1.ex}%
  \setlength{\abovedisplayskip}{1.ex} \setlength{\abovedisplayshortskip}{1.ex}%
  \begin{align*}
    \mathbb{E} {\Vert \vt_{\vlambda}\left(\rvvu\right) - \bar{\vzeta}_{\mathrm{H}} \rVert}_2^2
    \leq \frac{2}{\mu} \left( f_{\mathrm{H}}\left(\vt_{\vlambda}\left(\rvvu\right)\right) - f^*_{\mathrm{H}}\right).
  \end{align*}
  }%
  However, for the entropy-regularized form, this soon runs into a dead end since in
  {%
  \setlength{\belowdisplayskip}{1ex} \setlength{\belowdisplayshortskip}{1ex}%
  \setlength{\abovedisplayskip}{1ex} \setlength{\abovedisplayshortskip}{1ex}%
  \begin{align*}
    \mathbb{E} f_{\mathrm{H}}\left(\vt_{\vlambda}\left(\rvvu\right)\right) - f^*_{\mathrm{H}}
    &= F\left(\vlambda\right) - h\left(\vlambda\right) - f^* \\
    &= \left( F\left(\vlambda\right) - F^* \right) + \left(F^* - f^*\right) - h_{\mathrm{H}}\left(\vlambda\right),
  \end{align*}
  }%
  the negative entropy term \(h_{\mathrm{H}}\) is not bounded unless we rely on assumptions that need modifications to the BBVI algorithms. (\textit{e.g.}, bounded support, bounded domain).
  Fortunately, the following inequality cleverly side-steps this problem:
  {%
  \setlength{\belowdisplayskip}{1.5ex} \setlength{\belowdisplayshortskip}{1.5ex}%
  \setlength{\abovedisplayskip}{1.5ex} \setlength{\abovedisplayshortskip}{1.5ex}%
  \begin{align}
    \hspace{-1.0em}
    \mathbb{E} {\Vert \vt_{\vlambda}\left(\rvvu\right) - \bar{\vzeta}_{\mathrm{H}} \rVert}_2^2
    &\leq
    2\,\mathbb{E} {\lVert \vt_{\vlambda}\left(\rvvu\right) - \bar{\vzeta}_{\text{KL}} \rVert}_2^2
    +
    2 \, {\lVert \bar{\vzeta}_{\text{KL}} - \bar{\vzeta}_{\mathrm{H}} \rVert}_2^2,
    \label{eq:thm_upper_bound_parallel}
  \end{align}
  }%
  albeit at the cost of some looseness.
  By converting the entropy-regularized form into the KL-regularized form, the regularizer term becomes \(h_{\mathrm{KL}} = \DKL{q_{\vlambda}}{p} \geq 0\), which is bounded below by definition, unlike the entropic-regularizer \(h_{\mathrm{H}}\). 
  The proof completes by 
  \begin{enumerate}
  \vspace{-1ex}
    \setlength\itemsep{-.5ex}
    \item[\ding{185}] applying the quadratic growth assumption to relate the parameter distance with the function suboptimality gap, and 
    \item[\ding{186}] upper bounding the KL regularizer term.
  \end{enumerate}
  \vspace{-4ex}
\end{proofsketch}
\vspace{-2ex}
\begin{proofEnd}
  The proof uses the \(L_{\mathrm{H}}\)-smoothness of \(f_{\mathrm{H}}\) such that 
  \begin{align}
    \mathbb{E} \norm{
    \nabla f_{\mathrm{H}}\left(\vt_{\vlambda}\left(\rvvu\right)\right)
    }_2^2
    &=
    \mathbb{E} {\lVert
    \nabla f_{\mathrm{H}}\left(\vt_{\vlambda}\left(\rvvu\right)\right)
    -
    \nabla f_{\mathrm{H}}\left(\bar{\vzeta}_{\mathrm{H}}\right)
    \rVert}_2^2
    \nonumber
    \\
    &\leq
    L^2_{\mathrm{H}}
    \mathbb{E} {\lVert
    \vt_{\vlambda}\left(\rvvu\right)
    -
    \bar{\vzeta}_{\mathrm{H}}
    \rVert}_2^2,
    \label{eq:thm_upper_bound_smoothness}
  \end{align}
  where \(\bar{\vzeta}_{\mathrm{H}}\) is a stationary point of \(f_{\mathrm{H}}\) such that \(\nabla f_{\mathrm{H}}\left(\bar{\vzeta}_{\mathrm{H}}\right) = \mathbf{0}\).
  These steps have been previously used by \citet[Theorem 3]{domke_provable_2019} to prove the special case for the matrix square root parameterization.

  For the mean-field parameterization, we start from \cref{thm:general_variational_gradient_norm_bound} and apply~\cref{eq:thm_upper_bound_smoothness} as
  \begin{align}
    &\mathbb{E} \norm{
      \nabla_{\vlambda} f_{\mathrm{H}}\left(\vt_{\vlambda}\left(\rvvu\right)\right)
    }_2^2 
    \nonumber
    \\
    &\;\leq
    \mathbb{E} \norm{
      \nabla f_{\mathrm{H}}\left(\vt_{\vlambda}\left(\rvvu\right)\right)
    }_2^2 \left(1 + \norm{\mathbfsfit{U}}_{\mathrm{F}}\right)
    \nonumber
    \\
    &\;\leq
    L^2_{\mathrm{H}} \,
    \mathbb{E} {\lVert
    \vt_{\vlambda}\left(\rvvu\right)
    +
    \bar{\vzeta}_{\mathrm{H}}
    \rVert}_2^2 \left(1 + \norm{\mathbfsfit{U}}_{\mathrm{F}}\right),
    \nonumber
\shortintertext{applying \cref{thm:meanfield_u_identity},}
    &\;\leq
    L^2_{\mathrm{H}}
    \left( \kappa \sqrt{d} + \sqrt{\kappa d} + 1 \right) {\lVert \vm - \bar{\vzeta}_{\mathrm{H}} \rVert}_2^2
    \nonumber
    \\
    &\;\;\quad+
    L^2_{\mathrm{H}} \left(2 \kappa \sqrt{d} + 1\right) \norm{\mC}_{\mathrm{F}}^2,
    \nonumber
\shortintertext{and since the kurtosis satisfies \(\kappa \geq 1\) and thus \(\kappa \geq \sqrt{\kappa}\),}
    &\;\leq
    L^2_{\mathrm{H}}
    \left( 2 \kappa \sqrt{d} + 1 \right) \left( {\lVert \vm - \bar{\vzeta}_{\mathrm{H}} \rVert}_2^2
    + \norm{\mC}_{\mathrm{F}}^2 \right).
    \label{eq:thm1_meanfield}
  \end{align}

  Similarly, for the full-rank parameterizations, we start from \cref{thm:general_variational_gradient_norm_bound} and apply~\cref{eq:thm_upper_bound_smoothness} as
  \begin{alignat}{2}
    &\mathbb{E} \norm{
      \nabla_{\vlambda} f_{\mathrm{H}}\left(\vt_{\vlambda}\left(\rvvu\right)\right)
    }_2^2 
    \\
    &\;\leq\mathbb{E}\norm{\nabla f_{\mathrm{H}}\left(\vt_{\vlambda}\left(\rvvu\right)\right)}_2^2 \left(1 + \norm{\rvvu}_2^2 \right),
    \nonumber
    \\
    &\;\leq
    L^2_{\mathrm{H}}\,\mathbb{E}{\lVert \vt_{\vlambda}\left(\rvvu\right) - \bar{\vzeta}_{\mathrm{H}} \rVert}_2^2 \left(1 + \norm{\rvvu}_2^2 \right),
    \nonumber
\shortintertext{applying \cref{thm:reparam_u_identity},}
    &\;=
    L^2_{\mathrm{H}}\left(\left(d + 1\right) {\lVert \vm - \bar{\vzeta}_{\mathrm{H}} \rVert}_2^2 + \left(d + \kappa\right) \norm{\mC}_{\mathrm{F}}^2 \right),
    \nonumber
\shortintertext{and since the kurtosis satisfies \(\kappa \geq 1\),}
    &\;\leq
    L^2_{\mathrm{H}}\left(d + \kappa\right) \left( {\lVert \vm - \bar{\vzeta}_{\mathrm{H}} \rVert}_2^2 + \norm{\mC}_{\mathrm{F}}^2 \right).
    \label{eq:thm1_fullrank}
  \end{alignat}
  
  Both \cref{eq:thm1_meanfield,eq:thm1_fullrank} can now be denoted as
  \begin{alignat}{2}
    \mathbb{E}\norm{\nabla_{\vlambda} f_{\mathrm{H}}\left(\vt_{\vlambda}\left(\rvvu\right)\right)}_2^2 
    &\leq
    L^2_{\mathrm{H}}\, C\left(d, \kappa\right)  \left( {\lVert \vm - \bar{\vzeta}_{\mathrm{H}} \rVert}_2^2 + \norm{\mC}_{\mathrm{F}}^2 \right),
    \nonumber
\shortintertext{where by \cref{thm:reparam_quadratic},}
    &=
    L^2_{\mathrm{H}}\, C\left(d, \kappa\right) \mathbb{E} {\Vert \vt_{\vlambda}\left(\rvvu\right) - \bar{\vzeta}_{\mathrm{H}} \rVert}_2^2,
    \label{eq:thm1_parameter_suboptimality_unified}
  \end{alignat}
  and the constants are \(C\left(d, \kappa\right) = \kappa \sqrt{d} + 1\) for mean-field and \(C\left(d, \kappa\right) = d + \kappa\) for the full-rank parameterizations.

  As mentioned in the sketch, it is necessary to convert the entropy-regularized form into the KL-regularized form through the following inequality:
  {%
  \setlength{\belowdisplayskip}{1ex} \setlength{\belowdisplayshortskip}{1ex}%
  \setlength{\abovedisplayskip}{1ex} \setlength{\abovedisplayshortskip}{1ex}%
  \begin{alignat*}{2}
    \hspace{-0.5em}
    \mathbb{E} {\Vert \vt_{\vlambda}\left(\rvvu\right) - \bar{\vzeta}_{\mathrm{H}} \rVert}_2^2
    &\leq
    2\,\mathbb{E} {\lVert \vt_{\vlambda}\left(\rvvu\right) - \bar{\vzeta}_{\text{KL}} \rVert}_2^2
    +
    2 \, {\lVert \bar{\vzeta}_{\text{KL}} - \bar{\vzeta}_{\mathrm{H}} \rVert}_2^2.
  \end{alignat*}
  }%
  where \(\bar{\vzeta}_{\mathrm{KL}} = \Pi_{f_{\mathrm{KL}}}\left(\bar{\vzeta}_{\mathrm{H}}\right)\) is a projection of \(\bar{\vzeta}_{\mathrm{H}}\) to the set of minimizers of \(f_{\mathrm{KL}}\).
  Note that the KL-regularized form does not need to be tractable; only its existence suffices.
  We can now apply the quadratic growth assumption as
  {%
  \setlength{\belowdisplayskip}{1ex} \setlength{\belowdisplayshortskip}{1ex}%
  \setlength{\abovedisplayskip}{1ex} \setlength{\abovedisplayshortskip}{1ex}%
  \begin{alignat}{2}
    \hspace{-1em}
    \mathbb{E} {\lVert \vt_{\vlambda}\left(\rvvu\right) - \bar{\vzeta}_{\text{KL}} \rVert}_2^2
    \nonumber
    &\leq
    \frac{2}{\mu_{\mathrm{KL}}} \left( \mathbb{E} f_{\text{KL}}\left(\vt_{\vlambda}\left(\rvvu\right)\right) - f^*_{\text{KL}} \right)
    \nonumber
    \\
    &=
    \frac{2}{\mu_{\mathrm{KL}}} 
    \left( F\left(\vlambda\right) - h_{\mathrm{KL}}\left(\vlambda\right) - f^*_{\text{KL}} \right),
    \nonumber
\shortintertext{and since \(-h_{\mathrm{KL}}\left(\vlambda\right) = -\DKL{q_{\vlambda}}{p} \leq 0\) by definition,}
    &\leq
    \frac{2}{\mu_{\mathrm{KL}}} 
    \left( F\left(\vlambda\right) - f^*_{\text{KL}} \right)
    \label{eq:thm_upper_bound_kl_upper_bound}
    \\
    &=
    \frac{2}{\mu_{\mathrm{KL}}} \left(
    \left( F\left(\vlambda\right) - F^* \right) 
    + \left( F^* - f^*_{\text{KL}} \right)
    \right).
    \label{eq:thm_upper_bound_f_quadratic}
  \end{alignat}
  }%
  Combining \cref{eq:thm1_parameter_suboptimality_unified} with \cref{eq:thm_upper_bound_parallel},
  {
  \begin{align*}
    &\mathbb{E}\norm{\nabla_{\vlambda} f_{\mathrm{H}}\left(\vt_{\vlambda}\left(\rvvu\right)\right)}_2^2 
    \\
    &\;\leq
    2 \, L^2_{\mathrm{H}}\, C\left(d, \kappa\right) \mathbb{E} {\Vert \vt_{\vlambda}\left(\rvvu\right) - \bar{\vzeta}_{\mathrm{H}} \rVert}_2^2
    + 2 \, L^2_{\mathrm{H}}\, C\left(d, \kappa\right) {\lVert \bar{\vzeta}_{\text{KL}} - \bar{\vzeta}_{\mathrm{H}} \rVert}_2^2,
\shortintertext{and applying \cref{eq:thm_upper_bound_f_quadratic},}
    &\;\leq
    \frac{4 \, L^2_{\mathrm{H}}}{\mu_{\mathrm{KL}}} C\left(d, \kappa\right) \left(
    \left( F\left(\vlambda\right) - F^* \right) 
    +
    \left( F^* - f^*_{\text{KL}} \right)
    \right)
    \\
    &\;\quad+
    2 \, L^2_{\mathrm{H}}\, C\left(d, \kappa\right) {\lVert \bar{\vzeta}_{\text{KL}} - \bar{\vzeta}_{\mathrm{H}} \rVert}_2^2
  \end{align*}
  }%
  Plugging this into \cref{thm:gradient_variance_general_upper_bound} yields the result.
\end{proofEnd}


\vspace{1ex}
\begin{remark}\label{remark:ml_map}
  If the bijector \(\psi\) is an identity function, \(\vzeta_{\mathrm{KL}}\) and \(\vzeta_{\mathrm{H}}\) are the maximum likelihood (ML) and maximum a-posteriori (MAP) estimates, respectively.
  Thus, with enough datapoints, the term \( {\lVert \bar{\vzeta}_{\mathrm{KL}} - \bar{\vzeta}_{\mathrm{H}} \rVert}_2^2 \) will be negligible since the ML and MAP estimates will be close.
\end{remark}

\begin{remark}\label{remark:tighter}
  It is also possible the tighten the constants by a factor of two. 
  Instead of applying \cref{eq:thm_upper_bound_parallel}, we can use the inequality 
  {%
  \setlength{\belowdisplayskip}{1.ex} \setlength{\belowdisplayshortskip}{1.ex}%
  \setlength{\abovedisplayskip}{1.ex} \setlength{\abovedisplayshortskip}{1.ex}%
  \[
     {(a + b)}^2 \leq \left(1 + \delta^{2}\right) a^2 + \left(1 + \delta^{-2}\right) b^2,
  \]
  }%
  for some \(\delta > 0\).
  By setting \(\delta^2 = b = {\lVert\bar{\vzeta}_{\mathrm{KL}} - \bar{\vzeta}_{\mathrm{H}}\rVert}_2\), 
  {%
  \setlength{\belowdisplayskip}{1.ex} \setlength{\belowdisplayshortskip}{1.ex}%
  \setlength{\abovedisplayskip}{1.ex} \setlength{\abovedisplayshortskip}{1.ex}%
  \[
    \mathbb{E} {\Vert \vt_{\vlambda}\left(\rvvu\right) - \bar{\vzeta}_{\mathrm{H}} \rVert}_2^2
    \leq
    (1 + \delta^2)\,\mathbb{E} {\lVert \vt_{\vlambda}\left(\rvvu\right) - \bar{\vzeta}_{\text{KL}} \rVert}_2^2
    +
    \delta^4
    +
    \delta^2.
  \]
  }%
  Since \(\delta \approx 0\) as explained in \cref{remark:ml_map}, the constant in front of the first term is tightened almost by a factor of 2.
  However, the stated form is more convenient for theory since the first term does not depend on \({\lVert\bar{\vzeta}_{\mathrm{KL}} - \bar{\vzeta}_{\mathrm{H}}\rVert}_2\).
\end{remark}

\begin{remark}
  Let \(\kappa_{\mathrm{cond.}} = \nicefrac{L_{\mathrm{H}}}{\mu_{\mathrm{KL}}}\) be the \textit{condition number} of the problem.
  For the full-rank parameterizations, the variance is bounded as \(\mathcal{O}\left( L_{\mathrm{H}} \kappa_{\mathrm{cond.}} \left(d + \kappa\right) / M \right)\).
  The variance depends linearly on 
  \begin{enumerate}
    \vspace{-1ex}
    \setlength\itemsep{-1ex}
    \item[\ding{182}] the scaling of the problem \(L_{\mathrm{H}}\), 
    \item[\ding{183}] the conditioning of the problem \(\kappa_{\mathrm{cond.}}\),
    \item[\ding{184}] the dimensionality of the problem \(d\), and
    \item[\ding{185}] the tail properties of the variational family \(\kappa\),
    \vspace{-1ex}
  \end{enumerate}
  where the number of Monte Carlo samples \(M\) linearly reduces the variance.
\end{remark}

\vspace{-1.ex}
\paragraph{KL-Regularized Form}
We now prove an equivalent result for the KL-regularized form.
Here, we do not have to rely on~\cref{eq:thm_upper_bound_parallel} since we already start from \(f_{\mathrm{KL}}\), which results in better constants.

\begin{theoremEnd}[\theoremproofoption,category=upperboundtheoremklform]{theorem}\label{thm:gradient_upper_bound_kl}
  Let \(\rvvg_{M}\) be an \(M\)-sample estimator of the gradient of the ELBO in KL-regularized form (\cref{def:kl_form}). 
  Also, assume that
  \begin{itemize}[leftmargin=3em]
    \vspace{-1.5ex}
    \setlength\itemsep{0ex}
    \item \(f_{\mathrm{KL}}\) is \(L_{\mathrm{KL}}\)-smooth,
    \item \(f_{\mathrm{KL}}\) is \(\mu_{\mathrm{KL}}\)-quadratically growing,
    \vspace{-1.5ex}
  \end{itemize}
  and~\cref{assumption:q,assumption:phi_lipschitz} hold.
  Then, the gradient variance is bounded above as
  {\small%
  \setlength{\belowdisplayskip}{1ex} \setlength{\belowdisplayshortskip}{1ex}%
  \setlength{\abovedisplayskip}{1ex} \setlength{\abovedisplayshortskip}{1ex}%
  \begin{align*}
    \mathbb{E}\norm{\rvvg_{M}}_2^2
    &\leq
    \frac{2 L^2_{\mathrm{KL}}}{\mu_{\mathrm{KL}} M} C\left(d, \kappa\right) \left( F\left(\vlambda\right) - F^* \right)
    + \norm{ \nabla F\left(\vlambda\right) }_2^2
    \qquad\qquad
    \\
    &\quad+ \frac{2 L^2_{\mathrm{KL}}}{\mu_{\mathrm{KL}} M} C\left(d, \kappa\right) \left( F^* - f_{\mathrm{KL}}^*\right),
  \end{align*}
  }%
  where
  {\small%
  \setlength{\belowdisplayskip}{1ex} \setlength{\belowdisplayshortskip}{1ex}%
  \setlength{\abovedisplayskip}{1ex} \setlength{\abovedisplayshortskip}{1ex}%
  \begin{alignat*}{2}
    C\left(d, \kappa\right) &= 2 \kappa \sqrt{d} + 1 &&\;\text{for mean-field,} \\
    C\left(d, \kappa\right) &= d + \kappa          &&\;\text{for the Cholesky and matrix square root,}
  \end{alignat*}
  }%
  \(F^* = \inf_{\vlambda \in \mathbb{R}^p} F\left(\vlambda\right)\), and \(f_{\mathrm{KL}}^* = \inf_{\vzeta \in \mathbb{R}^d} f\left(\zeta\right)\).
\end{theoremEnd}
\vspace{-1ex}
\begin{proofEnd}
  This proof uses the smoothness of \(f_{\mathrm{KL}}\) instead of \(f_{\mathrm{H}}\).
  That is,
  \begin{align}
    \mathbb{E} \norm{
      \nabla f_{\mathrm{KL}}\left(\vt_{\vlambda}\left(\rvvu\right)\right)
    }_2^2 
    &=
    \mathbb{E} {\lVert
      \nabla f_{\mathrm{KL}}\left(\vt_{\vlambda}\left(\rvvu\right)\right)
      -
      \nabla f_{\mathrm{KL}}\left( \bar{\vzeta}_{\mathrm{KL}} \right)
    \rVert}_2^2 
    \nonumber
\shortintertext{applying \cref{eq:thm_upper_bound_smoothness},}
    &\leq
    L^2_{\mathrm{KL}} \,
    \mathbb{E} {\lVert
    \vt_{\vlambda}\left(\rvvu\right)
    -
    \bar{\vzeta}_{\mathrm{KL}}
    \rVert}_2^2, \label{eq:thm3_kl_smoothness}
  \end{align}
  where \(\bar{\vzeta}_{\mathrm{KL}}\) is a stationary point of \(f_{\mathrm{KL}}\).

  Substituting \cref{eq:thm3_kl_smoothness} in \cref{eq:thm1_parameter_suboptimality_unified},
  \begin{alignat}{2}
    &\mathbb{E}\norm{\nabla_{\vlambda} f_{\mathrm{KL}}\left(\vt_{\vlambda}\left(\rvvu\right)\right)}_2^2 
    \nonumber
    \\
    &\;\leq
    L^2_{\mathrm{KL}} C\left(d, \kappa\right) \mathbb{E} {\Vert \vt_{\vlambda}\left(\rvvu\right) - \bar{\vzeta}_{\mathrm{KL}} \rVert}_2^2,
    \nonumber
\shortintertext{and by applying \cref{eq:thm_upper_bound_f_quadratic} for \(f_{\mathrm{KL}}\),}
    &\;=
    \frac{2 L^2_{\mathrm{KL}} }{\mu_{\mathrm{KL}}} C\left(d, \kappa\right) \left( F\left(\vlambda\right) - F^*  \right) 
    + \frac{2 L^2_{\mathrm{KL}} }{\mu_{\mathrm{KL}}} C\left(d, \kappa\right)
    \left(
      F^* - f^*_{\text{KL}}
    \right).
    \nonumber
  \end{alignat}
  Plugging this to \cref{thm:gradient_variance_general_upper_bound} proves the result.
\end{proofEnd}


\vspace{-.5ex}
\subsection{Upper Bound Under Bounded Entropy}
\vspace{-.5ex}
The bound in \cref{thm:gradient_upper_bound} is slightly loose due to the use of~\cref{eq:thm_upper_bound_parallel} and \cref{eq:thm_upper_bound_kl_upper_bound}.
An alternative bound with slightly tighter constants, although the gains are marignal compared to \cref{remark:tighter}, can be obtained by assuming the following:
\begin{assumption}[\textbf{Bounded Entropy}]\label{assumption:bounded_entropy}
  The regularization term is bounded below as
  \( h_{\mathrm{H}}\left(\vlambda\right) \geq h_{\mathrm{H}}^* \).
\end{assumption}
For the entropy-regularized form, this corresponds to the entropy being bounded above by some constant since \(h\left(\vlambda\right) = - \mathrm{H}\left(q_{\vlambda}\right)\).
When using the nonlinear parameterizations (\cref{def:meanfield,def:fullrank}), this assumption can be practically enforced by bounding the output of \(\phi\) by some large \(S\).
\vspace{.5ex}
\begin{proposition}
    Let the diagonal conditioner \(\phi\) be bounded as \(\phi\left(x\right) \leq S\).
    Then, for any \(d\)-dimensional distribution \(q_{\vlambda}\) in the location-scale family with the mean-field (\cref{def:meanfield}) or Cholesky (\cref{def:fullrank}) parameterizations,
  {%
  \setlength{\belowdisplayskip}{1.ex} \setlength{\belowdisplayshortskip}{1.ex}%
  \setlength{\abovedisplayskip}{1.ex} \setlength{\abovedisplayshortskip}{1.ex}%
    \[ h_{\mathrm{H}}\left(\vlambda\right) = -\mathrm{H}\left(q_{\vlambda}\right) \geq -\mathrm{H}\left(\varphi\right) - \frac{d}{2} \log S. \]
  }%
\end{proposition}
\vspace{-2ex}
\begin{proof}
    From \cref{thm:location_scale_entropy}, \(\mathrm{H}\left(q_{\vlambda}\right) = \mathrm{H}\left(\varphi\right) + \log \abs{\mC} \).
    Since \(\mC\) under \cref{def:meanfield,def:fullrank} is a diagonal or triangular matrix, the log absolute determinant is the log sum of the diagonals.
    The conclusion follows from the fact that the diagonals \(C_{ii} = \phi\left(s_i\right)\) are bounded by \(S\).
\end{proof}
This is essentially a weaker version of the bounded domain assumption, though only the diagonal elements of \(\mC\), \(s_1, \ldots, s_d\), are bounded.
While this assumption results in an admittedly less realistic algorithm, it enables a tighter bound for the entropy-regularized form ELBO.

\begin{theoremEnd}[\theoremproofoption,category=upperboundtheoremboundedentropy]{theorem}\label{thm:gradient_upper_bound_bounded_entropy}
  Let \(\rvvg_{M}\) be an \(M\)-sample estimator of the gradient of the ELBO in entropy-regularized form (\cref{def:entropy_form}). 
  Also, assume that
  \begin{itemize}[leftmargin=3em]
    \vspace{-1.5ex}
    \setlength\itemsep{0ex}
    \item \(f_{\mathrm{H}}\) is \(L_{\mathrm{H}}\)-smooth,
    \item \(f_{\mathrm{H}}\) is \(\mu_{\mathrm{H}}\)-quadratically growing,
    \item \(h_{\mathrm{H}}\) is bounded as \(h_{\mathrm{H}}\left(\vlambda\right) > h_{\mathrm{H}}^*\) (\cref{assumption:bounded_entropy}),
      \vspace{-1.5ex}
  \end{itemize}
  and~\cref{assumption:q,assumption:phi_lipschitz} hold.
  Then, the gradient variance of \(\vg_{M}\) is bounded above as
  {\small%
  \setlength{\belowdisplayskip}{1.ex} \setlength{\belowdisplayshortskip}{1.ex}%
  \setlength{\abovedisplayskip}{1.ex} \setlength{\abovedisplayshortskip}{1.ex}%
  \begin{align*}
    \mathbb{E}\norm{\rvvg_{M}}_2^2
    &\leq
    \frac{2 L^2_{\mathrm{H}}}{\mu_{\mathrm{H}} M} C\left(d, \kappa\right) \left( F\left(\vlambda\right) - F^* \right)
    + \norm{ \nabla F\left(\vlambda\right) }_2^2
    \qquad\qquad
    \\
    &\quad+ \frac{2 L^2_{\mathrm{H}}}{\mu_{\mathrm{H}} M} C\left(d, \kappa\right) \left( F^* - f_{\mathrm{H}}^* - h^*_{\mathrm{H}} \right),
  \end{align*}
  }
  where
  {\small%
  \setlength{\belowdisplayskip}{1.ex} \setlength{\belowdisplayshortskip}{1.ex}%
  \setlength{\abovedisplayskip}{1.ex} \setlength{\abovedisplayshortskip}{1.ex}%
  \begin{alignat*}{2}
    C\left(d, \kappa\right) &= 2 \kappa \sqrt{d} + 1 &&\;\text{for mean-field,} \\
    C\left(d, \kappa\right) &= d + \kappa          &&\;\text{for the Cholesky parameterization,}
  \end{alignat*}
  }
  \(F^* = \inf_{\vlambda \in \mathbb{R}^p} F\left(\vlambda\right)\), and \(f_{\mathrm{H}}^* = \inf_{\vzeta \in \mathbb{R}^d} f\left(\zeta\right)\).
\end{theoremEnd}
\vspace{-2ex}
\begin{proofsketch}
   Instead of using \cref{eq:thm_upper_bound_parallel}, we apply the quadratic assumption directly to \(f_{\text{H}}\).
   The remaining entropic-regularizer term can now be bounded through the bounded entropy assumption.
\end{proofsketch}
\vspace{-2ex}
\begin{proofEnd}
  The proof is similar to that of \cref{thm:gradient_upper_bound}.
  As mentioned in the \textit{proof sketch}, we use the fact that the entropic regularizer is bounded such that
  \[ -h_{\mathrm{H}}\left(\vlambda\right) < -h_{\mathrm{H}}^*. \]
  By applying the quadratic growth assumption directly to \(f_{\mathrm{H}}\), 
  \begin{alignat}{2}
    \mathbb{E} {\lVert \vt_{\vlambda}\left(\rvvu\right) - \bar{\vzeta}_{\text{H}} \rVert}_2^2
    &\leq
    \frac{2}{\mu_{\mathrm{H}}} \left( \mathbb{E} f_{\text{H}}\left(\vt_{\vlambda}\left(\rvvu\right)\right) - f^*_{\text{H}} \right)
    \nonumber
    \\
    &=
    \frac{2}{\mu_{\mathrm{H}}} 
    \left( F\left(\vlambda\right) - h_{\mathrm{H}}\left(\vlambda\right) - f^*_{\text{H}} \right),
    \nonumber
\shortintertext{and by \cref{assumption:bounded_entropy},}
    &\leq
    \frac{2}{\mu_{\mathrm{H}}} 
    \left( F\left(\vlambda\right) - F^* \right) 
    + \frac{2}{\mu_{\mathrm{H}}} \left( F^* - f^*_{\text{H}} - h_{\mathrm{H}}^*\right).
    \label{eq:thm_upper_bound_f_H_quadratic}
  \end{alignat}
  
  The proof resumes from \cref{eq:thm1_parameter_suboptimality_unified} as
  {%
  \setlength{\belowdisplayskip}{1.ex} \setlength{\belowdisplayshortskip}{1.ex}%
  \setlength{\abovedisplayskip}{1.ex} \setlength{\abovedisplayshortskip}{1.ex}%
  \begin{alignat}{2}
    &\mathbb{E}\norm{\nabla_{\vlambda} f_{\mathrm{H}}\left(\vt_{\vlambda}\left(\rvvu\right)\right)}_2^2
    \nonumber
    \\
    &\;\leq
    L^2_{\mathrm{H}} C\left(d, \kappa\right) \mathbb{E} {\Vert \vt_{\vlambda}\left(\rvvu\right) - \bar{\vzeta}_{\mathrm{H}} \rVert}_2^2,
    \nonumber
\shortintertext{and by applying \cref{eq:thm_upper_bound_f_H_quadratic},}
    &\;=
    \frac{2 L^2_{\mathrm{H}} }{\mu_{\mathrm{H}}} C\left(d, \kappa\right) \left( F\left(\vlambda\right) - F^*  \right) 
    + \frac{2 L^2_{\mathrm{H}} }{\mu_{\mathrm{H}}} C\left(d, \kappa\right)
    \left(
      F^* - f^*_{\mathrm{H}} - h_{\mathrm{H}}^*
    \right).
    \nonumber
  \end{alignat}
  }%
  Plugging this to \cref{thm:gradient_variance_general_upper_bound} proves the result.
\end{proofEnd}


\begin{figure*}[t]
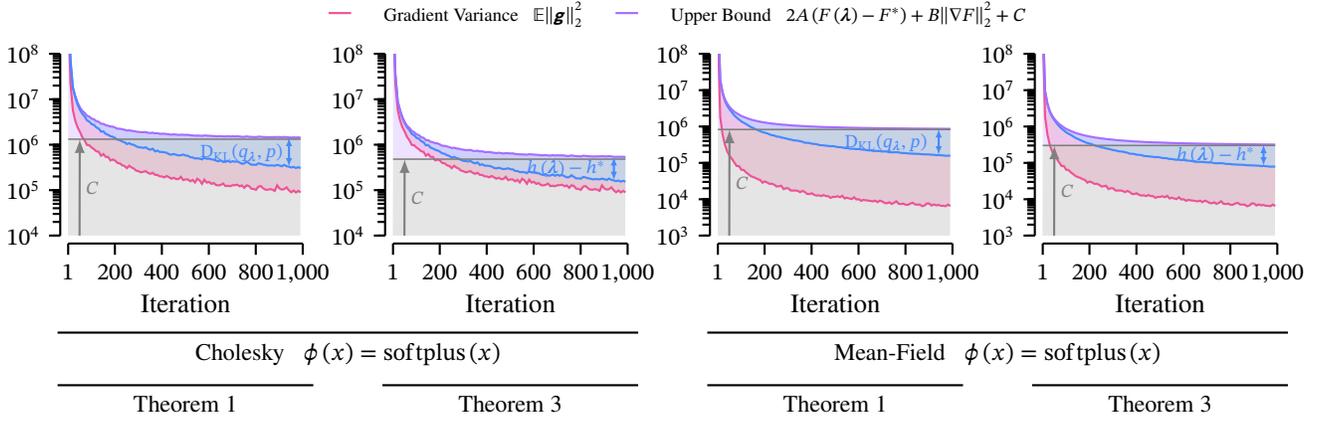

  \vspace{-2ex}
  \centering
{\hypersetup{linkbordercolor=black,linkcolor=black}
\begin{tikzpicture}
  \begin{groupplot}
    [
      group style={
        group size=4 by 1,
        horizontal sep=0.07\textwidth
      },
    ]
    \input{figures/group_quadratic_softpluschol_generalbound}
    \input{figures/group_quadratic_softpluschol_boundedentropy}
    \input{figures/group_quadratic_softplusmf_generalbound}
    \input{figures/group_quadratic_softplusmf_boundedentropy}
  \end{groupplot}
  
  \coordinate (c1r1southwest) at ($(group c1r1) + (-1.7cm, -2.5cm)$); 
  \coordinate (c1r1southeast) at ($(group c1r1) + (1.7cm,  -2.5cm)$); 

  \coordinate (c2r1southwest) at ($(group c2r1) + (-1.7cm, -2.5cm)$);
  \coordinate (c2r1southeast) at ($(group c2r1) + (1.7cm,  -2.5cm)$);

  \coordinate (c3r1southwest) at ($(group c3r1) + (-1.7cm, -2.5cm)$); 
  \coordinate (c3r1southeast) at ($(group c3r1) + ( 1.7cm, -2.5cm)$); 

  \coordinate (c4r1southwest) at ($(group c4r1) + (-1.7cm, -2.5cm)$);
  \coordinate (c4r1southeast) at ($(group c4r1) + ( 1.7cm, -2.5cm)$);
  
  \draw[thick,color=black] (c1r1southwest) -- (c2r1southeast) node[midway,below] {\small Cholesky\;\; \(\phi\left(x\right) = \mathrm{softplus}\left(x\right)\)};

  \draw[thick,color=black] (c3r1southwest) -- (c4r1southeast) node[midway,below] {\small Mean-Field\;\; \(\phi\left(x\right) = \mathrm{softplus}\left(x\right)\)};

  \draw[thick,color=black] ($(c1r1southwest) + (0,-0.7cm)$) -- ($(c1r1southeast) + (0,-0.7cm)$) node[midway,below] {\small\cref{thm:gradient_upper_bound}};
  \draw[thick,color=black] ($(c2r1southwest) + (0,-0.7cm)$) -- ($(c2r1southeast) + (0,-0.7cm)$) node[midway,below] {\small\cref{thm:gradient_upper_bound_bounded_entropy}};
  \draw[thick,color=black] ($(c3r1southwest) + (0,-0.7cm)$) -- ($(c3r1southeast) + (0,-0.7cm)$) node[midway,below] {\small\cref{thm:gradient_upper_bound}};
  \draw[thick,color=black] ($(c4r1southwest) + (0,-0.7cm)$) -- ($(c4r1southeast) + (0,-0.7cm)$) node[midway,below] {\small\cref{thm:gradient_upper_bound_bounded_entropy}};

  \node at ($(group c2r1) + (2.25cm,1.75cm)$) {\ref*{grouplegend}}; 
\end{tikzpicture}
}
  \vspace{-3ex}
  \caption{
    \textbf{Evaluation of the bounds for a perfectly conditioned quadratic target function.}
    The \textcolor{color3}{blue regions} are the loosenesses resulting from either using (\cref{thm:gradient_upper_bound}) or not using (\cref{thm:gradient_upper_bound_bounded_entropy}) the bounded entropy assumption (\cref{assumption:bounded_entropy}), while the \textcolor{color1}{red regions} are the remaining ``technical loosesnesses.''
    The gradient variance was estimated from \(10^3\) samples.
  }\label{fig:quadratic}
  \vspace{-2ex}
\end{figure*}





\vspace{-1ex}
\subsection{Matching Lower Bound}
\vspace{-.5ex}
Finally, we present a matching lower bound on the gradient variance of BBVI.
Our lower bound holds broadly for smooth and strongly convex problem instances that are well-conditioned and high-dimensional.

\vspace{.5ex}

\begin{theoremEnd}[\theoremproofoption,category=lowerboundtheorem]{theorem}\label{thm:gradient_lower_bound}
Let \(\rvvg_{M}\) be an \(M\)-sample estimator of the gradient of the ELBO in either the entropy- or KL-regularized form.
Also, let ~\cref{assumption:q} hold where the matrix square root parameterization is used.
Then, for all \(L\)-smooth and \(\mu\)-strongly convex functions \(f\) such that $\nicefrac{L}{\mu} < \sqrt{d + 1}$, the variance of \(\rvvg_{M}\) is bounded below by some strictly positive constant as
\begin{align*}
  \mathbb{E}\norm{\rvvg_M}_2^2
  &\geq
  \frac{2\mu^2 \left(d + 1\right) - 2 L^2}{ML} \left( F\left(\vlambda\right) - F^* \right) 
  + \norm{ \nabla F\left(\vlambda\right) }_2^2 \\
  &\quad+ \frac{2 \mu^2 \left(d + 1\right) - 2 L^2}{ML} \left(\mathbb{E}{f\left(\vt_{\vlambda^*}\left(\vu\right)\right)} - f^*\right),  
\end{align*}
as long as $\vlambda$ is in a local neighborhood around the unique global optimum $\vlambda^* = \argmin_{\vlambda \in \mathbb{R}^p} F\left(\vlambda\right)$, where \(F^* = F\left(\vlambda^*\right)\) and \(f^* = \argmin_{\vzeta \in \mathbb{R}^d} f\left(\vzeta\right)\).
\end{theoremEnd}
\vspace{-3ex}
\begin{proofsketch}
  We use the fact that, with the matrix square root parameterization, if \(f\) is \(L\)-smooth, $\mathbb{E} f\left(\vt_{\vlambda}\left(\rvvu\right)\right)$ is also $L$-smooth~\citep{domke_provable_2020}.
  From this, the parameter suboptimality can be related to the function suboptimality as
  {%
  \setlength{\belowdisplayskip}{1ex} \setlength{\belowdisplayshortskip}{1ex}%
  \setlength{\abovedisplayskip}{1ex} \setlength{\abovedisplayshortskip}{1ex}%
  \begin{align*}
    {\lVert \vlambda - \bar{\vlambda} \rVert}^2_2
    \geq
    \left({2}/{L}\right)
    \left(
    \mathbb{E}{f\left(\vt_{\vlambda}\left(\rvvu\right)\right)} 
    - f^*
    \right),
  \end{align*}
  }%
  where \(\bar{\vlambda} = \left(\bar{\vzeta}, \boldupright{O}\right)\).
  For the entropy term, we circumvent the need to directly bound its value by restricting our interest to the neighborhood of the minimizer \(\vlambda^*\), where the contribution of \(h\left(\vlambda^*\right) - h\left(\vlambda\right)\) will be marginal enough for the lower bound to hold.
\end{proofsketch}

\vspace{-2ex}
\begin{proofEnd}
When using the matrix square root parameterization,~\citet{domke_provable_2020} have shown that if $f$ is $L$-smooth, $\mathbb{E} f\left(\vt_{\vlambda}\left(\rvvu\right)\right)$ is also $L$-smooth.
Therefore, we have
\begin{align} 
    \norm{\mathbb{E}{ \nabla_{\vlambda} f\left(\vt_{\vlambda}\left(\rvvu\right)\right)} }_2^2 \leq 2 L \left(\mathbb{E} f\left(\vt_{\vlambda}\left(\rvvu\right)\right) - f^*\right).
    \label{eq:thm_lower_bound_eq1}
\end{align}

Furthermore, let $\bar{\vzeta}$ be the minimizer of $f$, namely $f^* = f\left(\bar{\vzeta}\right)$.
From \cref{thm:variational_gradient_norm_identity}, we have
\begin{align*}
    \mathbb{E}{\norm{\nabla_{\vlambda} f\left(\vt_{\vlambda}\left(\rvvu\right)\right)}_2^2} 
    & = \mathbb{E}{\norm{\nabla f\left(\vt_{\vlambda}\left(\rvvu\right)\right)}_2^2 \left(1 + \norm{\rvvu}_2^2\right)},
\shortintertext{by the \(\mu\)-strong convexity of \(f\),}
    & \geq 2 \mu \, \mathbb{E}{\left(f\left(\vt_{\vlambda}\left(\rvvu\right)\right) - f^*\right) \left(1 + \norm{\rvvu}_2^2\right)} \\
    & \geq \mu^2 \, \mathbb{E}{{\lVert\vt_{\vlambda}\left(\rvvu\right) - \bar{\vzeta}\rVert}^2_2 \left(1 + \norm{\rvvu}_2^2\right)},
\shortintertext{applying \Cref{thm:reparam_u_identity},}
    & = \mu^2 \, \left(\left(d + 1\right) {\lVert \vm - \bar{\vzeta} \rVert}^2_2 + \left(d + \kappa\right) \norm{\mC}_{\mathrm{F}}^2\right),
\shortintertext{and by the property of the kurtosis that \(\kappa \geq 1\),}
    & \geq \mu^2 \, \left(d + 1\right) {\lVert \vlambda - \bar{\vlambda} \rVert}^2_2,
\end{align*}
where $\bar{\vlambda} = \left(\bar{\vzeta}, \boldupright{O}\right)$.

Observe that $\bar{\vlambda}$ is the minimizer of $\mathbb{E}{f\left(\vt_{\vlambda}\left(\rvvu\right)\right)}$ 
 such that 
\[
   \mathbb{E}{f\left(\vt_{\bar{\vlambda}}\left(\rvvu\right)\right)} = f\left(\bar{\vzeta}\right) = f^* \leq \mathbb{E}{f\left(\vt_{\vlambda}\left(\rvvu\right)\right)}
\]
for any $\vlambda$.
Furthermore, from the $L$-smoothness of $\mathbb{E}{f\left(\vt_{\vlambda}\left(\rvvu\right)\right)}$, we have
\begin{align*}
    &\mu^2 \left(d + 1\right) {\lVert \vlambda - \bar{\vlambda} \rVert}^2_2 \\
    &\quad\geq \frac{2 \mu^2 \left(d + 1\right)}{L} \left(\mathbb{E}{f\left(\vt_{\vlambda}\left(\rvvu\right)\right)} - \mathbb{E}{f\left(\vt_{\bar{\vlambda}}\left(\rvvu\right)\right)}\right).
\end{align*}
Thus, we have
\begin{align}
    \mathbb{E}{\norm{\nabla_{\vlambda} f\left(\vt_{\vlambda}\left(\rvvu\right)\right)}_2^2} 
    &\geq 
    \frac{2 \mu^2 \left(d + 1\right)}{L} \left(\mathbb{E}{f\left(\vt_{\vlambda}\left(\rvvu\right)\right)} - f^*\right).
    \label{eq:thm_lower_bound_eq2}
\end{align}

Now, from \cref{eq:thm_gradient_variance_general_definition},
\begin{align*}
\mathbb{E}\norm{ \vg_M }_2^2
  &= \frac{1}{M} \left(
        \mathbb{E}{ \norm{\nabla_{\vlambda} f\left(\vt_{\vlambda}\left(\rvvu\right)\right)}_2^2 }
        -
        \norm{\mathbb{E}{ \nabla_{\vlambda} f\left(\vt_{\vlambda}\left(\rvvu\right)\right)} }_2^2
    \right) \\
    &\qquad+ \norm{ \nabla F\left(\vlambda\right) }^2_2,
\shortintertext{applying \cref{eq:thm_lower_bound_eq1},}
  &\geq
  \frac{1}{M} 
  \left(
    \mathbb{E}{ \norm{\nabla_{\vlambda} f\left(\vt_{\vlambda}\left(\rvvu\right)\right)}_2^2 }
    -
    2 L^2 \left(\mathbb{E}{f\left(\vt_{\vlambda}\left(\rvvu\right)\right)} - f^*\right) 
  \right) \\
  &\qquad+ \norm{ \nabla F\left(\vlambda\right) }^2_2 
\shortintertext{applying \cref{eq:thm_lower_bound_eq2},}
  &\geq
  \frac{2 \mu^2 \left(d + 1\right) - 2L^2}{ML} 
  \left(\mathbb{E}{f\left(\vt_{\vlambda}\left(\rvvu\right)\right)} - f^*\right) \\
  &\qquad + \norm{ \nabla F\left(\vlambda\right) }^2_2 
  \\
  &\geq 
  \frac{2 \mu^2 \left(d + 1\right) - 2L^2}{ML} \left(F\left(\vlambda\right) - h\left(\vlambda\right) - f^*\right) \\
  &\qquad+ \norm{ \nabla F\left(\vlambda\right) }^2_2 
  \\
  &= \frac{2 \mu^2 \left(d + 1\right) - 2L^2}{ML} \left(F\left(\vlambda\right) - F^*\right) + \norm{ \nabla F\left(\vlambda\right) }^2_2  \\
  &\qquad+ \frac{2 \mu^2 \left(d + 1\right) - 2 L^2}{ML} \left(F^* - f^* - h\left(\vlambda\right) \right).
\end{align*}
The last term
\begin{align*}
    \frac{2 \mu^2 \left(d + 1\right) - 2 L^2}{ML} \left(F^* - f^* - h\left(\vlambda\right) \right)
\end{align*}
can be shown to be positive if $\vlambda$ is sufficiently close to the optimum.
Let $\vlambda^* = \argmin_{\vlambda} F\left(\vlambda\right)$ be the minimizer of $F$.
Then, we have
\begin{align*}
    F^* - f^* - h\left(\vlambda\right)  
    &= 
    \mathbb{E}{f\left(\vt_{\vlambda^*}\left(\vu\right)\right)} + h\left(\vlambda^*\right) - f^* - h\left(\vlambda\right) 
    \\
    &=
    \left(\mathbb{E}{f\left(\vt_{\vlambda^*}\left(\vu\right)\right)} - f^*\right) + \left(h\left(\vlambda^*\right) - h\left(\vlambda\right)\right),
\end{align*}
where the first term is strictly positive and the second term goes to zero as $\vlambda \to \vlambda^*$.
\end{proofEnd}


\begin{remark}[\textbf{Matching Dimensional Dependence}]
  For well-conditioned problems such that \(\nicefrac{L}{\mu} < \sqrt{d+1}\), a lower bound of the same dimensional dependence with our upper bounds holds near the optimum. 
\end{remark}

\begin{remark}[\textbf{Unimprovability of the ABC Condition}]
  The lower bound suggests that the \(ABC\) gradient variance condition is unimprovable within the class of smooth, quadratically growing functions.
\end{remark}



\vspace{-1ex}
\section{Simulations}
\vspace{-.5ex}
We now evaluate our bounds and the insights gathered during the analysis through simulations.
We implemented a bare-bones implementation of BBVI in Julia~\citep{bezanson_julia_2017} with plain SGD.
The stepsize were manually tuned so that all problems converge at similar speeds.
For all problems, we use a unit Gaussian base distribution such that \(\varphi\left(u\right) = \mathcal{N}\left(u; 0, 1\right)\) resulting in a kurtosis of \(\kappa = 3\) and use \(M = 10\) Monte Carlo samples.






\vspace{-.5ex}
\subsection{Synthetic Problem}\label{section:quadratic}
To test the \textit{ideal} tightness of the bounds, we consider quadratics achieving the tightest bound for the constants \(L_{\mathrm{H}}, L_{\mathrm{KL}}, \mu_{\mathrm{H}}, \mu_{\mathrm{KL}}\) given as
{%
\setlength{\belowdisplayskip}{1.ex} \setlength{\belowdisplayshortskip}{1.ex}%
\setlength{\abovedisplayskip}{1.ex} \setlength{\abovedisplayshortskip}{1.ex}%
\[
  \log \ell\left(\vx \mid \vz \right) = -\frac{N}{\sigma^2} \norm{ \vz - \vz^* }_2^2;\quad
  \log p\left(\vz \right)             = -\frac{1}{\lambda}  \norm{ \vz  }_2^2,
\]
}%
where \(N\) simulates the effect of the number of datapoints.
We set the constants as \(\sigma = 0.3\), \(\lambda = 8.0\), and \(N = 100\), the mode \(\vz^*\) is randomly sampled from a Gaussian, and the dimension of the problem is \(d = 20\).
For the bounded entropy case, we set \(S = 2.0\) (the true standard deviation is in the order of 1e-3).

\vspace{-.5ex}
\paragraph{Quality of Upper Bound}
The results for the Cholesky and mean-field parameterizations with a softplus bijector are shown in~\cref{fig:quadratic}.
For the Cholesky parameterization, the bulk of the looseness comes from the treatment of the regularization term (\textcolor{color3}{blue region}).
The remaining ``technical looseness'' (\textcolor{color1}{red region}) is relatively tight and can be shown to be tighter when using linear parameterizations (\(\phi\left(x\right) = x\)) and the square root parameterization, which is the tightest.
However, for the mean-field parameterization, despite the superior constants~(\cref{remark:meanfield_superiority}), there is still room for improvement.
Additional results for other parameterizations can be found in~\cref{section:additional_quadratic}.

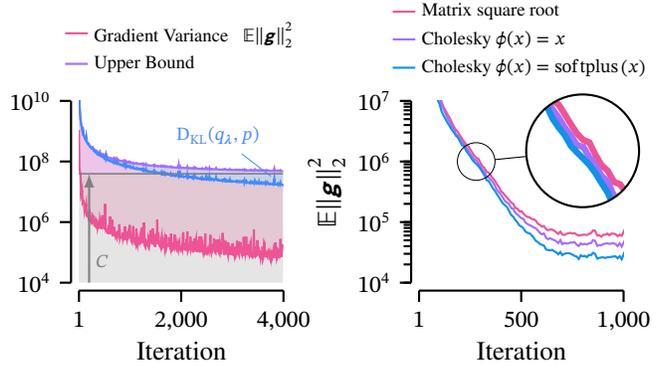
\begin{figure}[t]
  \vspace{-1.5ex}
  \hspace{-1.5em}
  \subfloat{
    \begin{tikzpicture}
\begin{axis}[
      legend style={
        legend image post style = {scale=0.5},
        at                      = {(0.5,1.5)},
        anchor                  = north,
        legend cell align       = left,
        line width              = 0.8pt,
        draw = none,
      },
      tuftelike, 
      xlabel style={yshift=-0.8ex},
      axis line style = thick,
      every tick/.style={black,thick},
      ymode  = log,
      xmin   = 1,
      xmax   = 4000,
      xtick  = {1, 2000, 4000},
      ytick  = {1e+4, 1e+6, 1e+8, 1e+10},
      ymin   = 1e+4,
      ymax   = 1e+10,
      xlabel = {Iteration},
      tick label style={font=\small},  
      height = 4cm,
      width  = 4.3cm,
      axis on top,
    ]
    \addplot[name path=gvar, color1, mark=none, thick]
      table [x=t, y=gvar] {data/simulation/airfoil_softpluschol_generalbound.csv}
      coordinate [pos=0.95] (gvarcoord);
    \addlegendentry{\scriptsize Gradient Variance\;\; \( \mathbb{E}\norm{ \rvvg }_2^2 \)}

    \addplot[name path=ABC, color2, mark=none, thick]
      table [x=t, y=ABC] {data/simulation/airfoil_softpluschol_generalbound.csv}
      coordinate [pos=0.05] (ABCcoord1) coordinate [pos=0.95] (ABCcoord2);
    \addlegendentry{\scriptsize Upper Bound}


    \addplot[name path=opt, color3, mark=none, thick] %
      table [x=t, y=opt] {data/simulation/airfoil_softpluschol_generalbound.csv} %
      coordinate [pos=0.95] (optcoord);

    \addplot[name path=axis, domain=0:3990, fill=none, no markers, draw=none] {1e+4}
      coordinate [pos=0.05] (axiscoord);


    \addplot[name path=C, fill=none, gray, mark=none]
      table [x=t, y=C] {data/simulation/airfoil_softpluschol_generalbound.csv}
      coordinate [pos=0.05] (Ccoord);

    \addplot[thick, color=color3, fill=color3, fill opacity=0.2] fill between[of=ABC and opt];
    \addplot[thick, color=color2, fill=color2, fill opacity=0.2] fill between[of=ABC and   C];

    \addplot[thick, color=color1, fill=color1, fill opacity=0.2] fill between[of=opt and gvar];
    \addplot[thick, color=gray,   fill=gray,   fill opacity=0.2] fill between[of=C   and axis];

    \draw[thick,color=color3,draw=none] (ABCcoord2) -- (optcoord)
      node[midway,xshift=-18pt] {}
      coordinate [pos=0.5] (DKLcoord);


    \draw[thick,color=gray,latex-] (Ccoord) -- (axiscoord)
      node[pos=0.8,xshift=5pt] {\scriptsize\(C\) };

    \node[%
      pin={%
        [%
          pin distance=0.3cm,%
          pin edge={color3},%
          text=color3%
        ]100:{\scriptsize\(\DKL{q_{\vlambda}}{p}\)}},%
      inner sep=0pt,%
    ] at (DKLcoord) {};
\end{axis}
\end{tikzpicture}\label{fig:airfoil_bound}
  }
  \subfloat{
    \hspace{-2em}
    \tikzset{subcaptionstyle/.style={
    text width=2in,yshift=-3mm, align=center,anchor=north
}}

\begin{tikzpicture}[%
    /pgfplots/set layers,
    spy using outlines={circle, magnification=3, connect spies}%
  ]
  \begin{axis}[
      legend style={
        legend image post style={scale=0.5},
        at={(0.5,1.6)},
        anchor=north,
        legend cell align=left,
        line width=0.8pt,
        draw=none 
      },
      tuftelike, 
      axis line style = thick,
      every tick/.style={black,thick},
      ymode  = log,
      xmin   = 1,
      xmax   = 1000,
      xtick  = {1, 500, 1000},
      xlabel style={yshift=-0.8ex},
      ymin   = 1e+4,
      ymax   = 1e+7,
      xlabel = {Iteration},
      tick label style={font=\small},  
      ylabel = {\(\mathbb{E}\norm{\rvvg}^2_2\)},
      height = 4cm,
      width  = 4.3cm,
      axis on top,
    ]
    \addplot[color1, thick, mark=none] table[x=t, y=squareroot, mark=none] {data/simulation/replay_airfoil.csv};
    \addlegendentry{\scriptsize{Matrix square root}};

    \addplot[color2, thick, mark=none] table[x=t, y=linearcholesky, mark=none] {data/simulation/replay_airfoil.csv};
    \addlegendentry{\scriptsize{Cholesky \(\phi(x) = x\)}};


    \addplot[color4, thick, mark=none] table[x=t, y=softpluscholesky, mark=none] {data/simulation/replay_airfoil.csv};
    \addlegendentry{\scriptsize{Cholesky \(\phi(x) = \mathrm{softplus}\left(x\right)\)}};

    \addplot [name path=squarerootl, fill=none, draw=none, forget plot] table [x=t, y=squarerootl] {data/simulation/replay_airfoil.csv} \closedcycle;
    \addplot [name path=squarerootu, fill=none, draw=none, forget plot] table [x=t, y=squarerootu] {data/simulation/replay_airfoil.csv} \closedcycle;
    \addplot[color1!30] fill between[of=squarerootu and squarerootl];

    \addplot [name path=linearcholeskyl, fill=none, draw=none, forget plot] table [x=t, y=linearcholeskyl] {data/simulation/replay_airfoil.csv} \closedcycle;
    \addplot [name path=linearcholeskyu, fill=none, draw=none, forget plot] table [x=t, y=linearcholeskyu] {data/simulation/replay_airfoil.csv} \closedcycle;
    \addplot[color2!40] fill between[of=linearcholeskyu and linearcholeskyl];


    \addplot [name path=softpluscholeskyl, fill=none, draw=none, forget plot] table [x=t, y=softpluscholeskyl] {data/simulation/replay_airfoil.csv} \closedcycle;
    \addplot [name path=softpluscholeskyu, fill=none, draw=none, forget plot] table [x=t, y=softpluscholeskyu] {data/simulation/replay_airfoil.csv} \closedcycle;
    \addplot[color4!40] fill between[of=softpluscholeskyu and softpluscholeskyl];

    \coordinate (spypoint)     at (axis cs:280,1e+6);
    \coordinate (magnifyglass) at (axis cs:800,1.5e+6);
    \begin{scope}
      \spy [black, size=1.5cm] on (spypoint) in node[fill=white] at (magnifyglass);
    \end{scope}
  \end{axis}
\end{tikzpicture}\label{fig:airfoil_parameterizations}
  }
  \vspace{-1.0ex}
  \caption{
    \textbf{
      Linear regression on the \textsc{Airfoil} dataset.
      (\textsf{left}) Evaluation of the upper bound (\cref{thm:gradient_upper_bound}).
      (\textsf{right}) Comparison of the variance of different parameterizations resulting in the same \(\vm\), \(\mC\).
    }
  }
  \vspace{-3.0ex}
\end{figure}

\vspace{-.5ex}
\subsection{Real Dataset}\label{section:linearreg}
\vspace{-.5ex}
\paragraph{Model}
We now evaluate the theoretical results with real datasets.
Given a regression dataset \((\mX, \vy)\), we use the linear Gaussian model 
{%
\setlength{\belowdisplayskip}{1.ex} \setlength{\belowdisplayshortskip}{1.ex}%
\setlength{\abovedisplayskip}{1.ex} \setlength{\abovedisplayshortskip}{1.ex}%
\[
  y   \sim \mathcal{N}\left(\mX \vw, \sigma^2\right);\quad
  \vw \sim \mathcal{N}\left(\mathbf{0}, \lambda \boldupright{I}\right),
\]
}%
where \(\lambda\) and \(\sigma\) are hyperparameters.
The smoothness and quadratic growth constants for this model are given as the max- and minimum eigenvalues of \(\sigma^{-2} \mX^{\top} \mX + \lambda^{-1} \boldupright{I}\) (for \(f_{\text{H}}\)) and \(\sigma^{-2} \mX^{\top} \mX\) (for \(f_{\text{KL}}\)).
\(f_{\text{KL}}^*\) and \(f_{\text{H}}^*\) are given as the mode of the likelihood and the posterior, while \(F^*\) is the negative marginal log-likelihood.

\vspace{-.5ex}
\paragraph{Quality of Upper Bound}
\cref{fig:airfoil_bound} shows the result on the \textsc{Airfoil} dataset~\citep{dua_uci_2017}.
The constants are \(L_{\mathrm{H}} = 3.520 \times 10^4, \mu_{\mathrm{KL}}=2.909 \times 10^3\).
Due to poor conditioning, the bound is much looser compared to the quadratic case.
We note that generalizing our bounds to utilize matrix smoothness and matrix-quadratic growth as done by \citep{domke_provable_2019} would tighten the bounds.
But the theoretical gains would be marginal.
Detailed information about the datasets and additional results for other parameterizations can be found in~\cref{section:additional_linearreg}.

\vspace{-.5ex}
\paragraph{Comparison of Parameterizations}
\cref{fig:airfoil_parameterizations} compares the gradient variance resulting from the different parameterizations.
For a fair comparison, the gradient is estimated on the \(\vlambda\) that results in the same \(\vm, \mC\) for all three parameterizations.
This shows the gradual increase in variance by
\begin{enumerate*}[label=\textbf{(\roman*)}]
  \item not using a nonlinear conditioner (linear Cholesky)
  \item and increasing the number of variational parameters (matrix square root).
\end{enumerate*}


\section{Related Works}
\vspace{-.5ex}
\paragraph{Controlling Gradient Variance}
The main algorithmic challenge in BBVI is to control the gradient noise~\cite{ranganath_black_2014}.
This has led to various methods for reducing the variance of VI gradient estimators using control variates~\citep{ranganath_black_2014,miller_reducing_2017,geffner_using_2018}, ensembling of estimators~\citep{geffner_rule_2020}, modifying the differentiation procedure~\citep{roeder_sticking_2017}, quasi-Monte Carlo~\citep{buchholz_quasimonte_2018, liu_quasimonte_2021}, and multilevel Monte Carlo~\citep{fujisawa_multilevel_2021}.
Cultivating a deeper understanding of the properties of gradient variance could further extend this list.

\vspace{-2ex}
\paragraph{Convergence Guarantees}
Obtaining full convergence guarantees has been an important task for understanding BBVI algorithms.
However, most guarantees so far have relied on strong assumptions such as that the log-likelihood is Lipschitz~\citep{cherief-abdellatif_generalization_2019,alquier_nonexponentially_2021}, that the gradient variance is bounded by constant~\citep{liu_quasimonte_2021,buchholz_quasimonte_2018,domke_provable_2020,hoffman_blackbox_2020}, and that the support of \(q_{\vlambda}\) is bounded~\citep{fujisawa_multilevel_2021}.
Our result shows that similar results can be obtained under relaxed assumptions.
Meanwhile,~\citet{bhatia_statistical_2022} have recently proven a full complexity guarantee for a variant of BBVI.
But similarly to \citet{hoffman_blackbox_2020}, they only optimize the scale matrix \(\mC\), and the specifics of the algorithm diverge from the usual BBVI implementations as it uses the stochastic power iterations instead of SGD.

\vspace{-1ex}
\paragraph{Gradient Variance Guarantees}
Studying the actual gradient variance properties of BBVI has only started to make progress recently.
~\citet{fan_fast_2015} first provided bounds by assuming the log-likelihood to be Lipschitz.
Under more general conditions,~\citet{domke_provable_2019} provided tight bounds for smooth log-likelihoods, which our work builds upon.
\citeauthor{domke_provable_2019}'s result can also be seen as a direct generalization of the results of~\citet{xu_variance_2019}, which are restricted to quadratic log-likelihoods and the mean-field family.
Lastly,~\citet{mohamed_monte_2020} provides a conceptual evaluation of gradient estimators used in BBVI.


\vspace{-.5ex}
\section{Discussions}
\vspace{-.5ex}
\paragraph{Conclusions}
In this work, we have proven upper bounds on the gradient variance of BBVI with the location-scale family for smooth, quadratically-growing log-likelihoods.
Specifically, we have provided bounds for both the ELBO in entropy-regularized and KL-regularized forms.
Our guarantees work without a single modification to the algorithms used in practice, although stronger assumptions establish a tighter bound for the entropy-regularized form ELBO.
Also, our bounds corresponds to the ABC condition (\cref{section:abc}) and the \textit{expected residual} (ER) condition, where the latter is a special case of the former with \(B=1\).
The ER condition has been used by \citet{gower_sgd_2021} for proving convergence of SGD on quasar convex functions, which generalize convex functions.
The results of this paper are used by~\citet{kim_blackbox_2023} to establish convergence of BBVI through the results of~\citet{khaled_better_2023}.

\vspace{-1ex}
\paragraph{Limitations}
Our results have the following limitations:
\begin{enumerate*}
    \item[\ding{182}] Our results only apply to smooth and quadratically-growing log likelihoods and
    \item[\ding{183}] the location-scale ADVI family. Also, 
    \item[\ding{184}] our bounds cannot distinguish the variance of the Cholesky and matrix square root parameterizations, 
    \item[\ding{185}] and empirically, the bounds for the mean-field parameterization appear loose. Furthermore, 
    \item[\ding{186}] our results only work with 1-Lipschitz diagonal conditioners such as the softplus function.
\end{enumerate*}
Unfortunately, assuming both smoothness and quadratic growth is quite restrictive, as it leaves a very small number of known distributions.
Also, in practice, non-Lipschitz conditioners such as the exponential functions are widely used.
While obtaining similar bounds with such conditioners would be challenging, constructing a theoretical framework that extends to such would be an important future research direction.

\vspace{-1ex}
\section*{Acknowledgements}
\vspace{-1ex}
This work was supported by NSF award IIS-2145644.

\clearpage
\bibliography{references}
\bibliographystyle{icml2023}

\clearpage
\appendix
\onecolumn



\clearpage

{\hypersetup{linkbordercolor=black,linkcolor=black}
\tableofcontents
}

{\hypersetup{linkbordercolor=black,linkcolor=black}
\section{Detailed Comparison Against \citealt*{domke_provable_2019}}
}%
Under the assumption that \(f_{\mathrm{H}}\) is \(L_{\mathrm{H}}\)-smooth and the linear full-rank Cholesky parameterization, under our notation, \citep{domke_provable_2019} prove the following bound:
\begin{equation*}
  \mathbb{E} \norm{\rvvg_{M=1}}_2^2
  \leq 
  L_{\mathrm{H}}^2 \left( (d+1) {\lVert \vm - \bar{\vzeta}_{\mathrm{H}} \rVert}_2^2 + (d+\kappa) \norm{\mC}_{\mathrm{F}}^2 \right).
  \label{eq:domke_bound}
\end{equation*}
We extend \citeauthor{domke_provable_2019}'s analysis in three original directions.

\paragraph{\ding{182} Generalization to Nonlinear Parameterizations}
First, we generalize the bounds to support nonlinear parameterizations.
In particular, \cref{thm:general_variational_gradient_norm_identity} and \cref{thm:general_variational_gradient_norm_bound} generalize Lemma 1 of \citet{domke_provable_2019} to 1-Lipschitz nonlinear conditioners.
From here, the analysis becomes identical to \citeauthor{domke_provable_2019}'s setup, until we reach our original analysis we discuss in Item \ding{184}.

\paragraph{\ding{183} Tighter Bound for the Mean-Field Parameterization}
Second, for the mean-field parameterization, we prove a bound that is tighter in the large \(d\) regime, 
\[
  \mathbb{E} \norm{\rvvg_{M=1}}_2^2
  \leq 
  L_{\mathrm{H}}^2 \left( ( \sqrt{d \kappa} + \kappa \sqrt{d} + 1 ) {\lVert \vm - \bar{\vzeta}_{\mathrm{H}} \rVert}_2^2 + (2 \kappa \sqrt{d} + 1) \norm{\mC}_{\mathrm{F}}^2 \right),
\]
as a direct consequence of \cref{thm:u_identities}.

\paragraph{\ding{184} Connecting with the ABC Condition}
Furthermore, we extend the bounds above and establish the ABC condition (\cref{assumption:abc}) for the ELBO, through the quadratic function growth condition (\cref{def:quadratic_growth}).
Specifically, in our proof of \cref{thm:gradient_upper_bound}, the derivation past \cref{eq:thm1_fullrank} is original.

\clearpage

\section{Additional Simulation Results}

\vspace{-1ex}
\subsection{Synthetic Problem}\label{section:additional_quadratic}
We provide additional results for the simulations with quadratics in~\cref{section:quadratic}.

\begin{figure}[H]
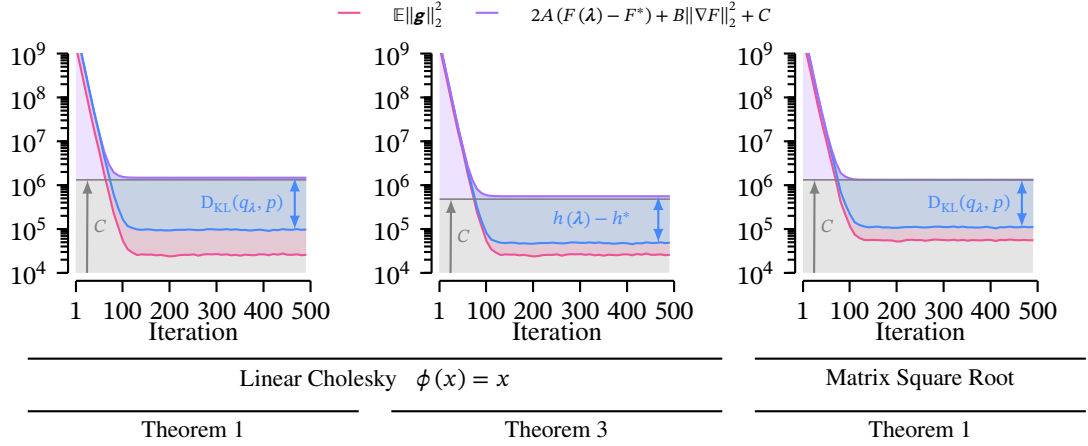

  \vspace{-2ex}
  \centering
{\hypersetup{linkbordercolor=black,linkcolor=black}
\begin{tikzpicture}
  \begin{groupplot}
    [
      group style={
        group size=3 by 1,
        horizontal sep=0.1\textwidth
      },
    ]
    \input{figures/group_quadratic_linearchol_generalbound}
    \input{figures/group_quadratic_linearchol_boundedentropy}
    \input{figures/group_quadratic_squareroot_generalbound}
  \end{groupplot}
  
  \coordinate (c1r1southwest) at ($(group c1r1) + (-2.2cm, -2.6cm)$); 
  \coordinate (c1r1southeast) at ($(group c1r1) + ( 2.2cm, -2.6cm)$); 

  \coordinate (c2r1southwest) at ($(group c2r1) + (-2.2cm, -2.6cm)$);
  \coordinate (c2r1southeast) at ($(group c2r1) + ( 2.2cm, -2.6cm)$);

  \coordinate (c3r1southwest) at ($(group c3r1) + (-2.2cm, -2.6cm)$); 
  \coordinate (c3r1southeast) at ($(group c3r1) + ( 2.2cm, -2.6cm)$); 

  \draw[thick,color=black] (c1r1southwest) -- (c2r1southeast) node[midway,below] {\small Linear Cholesky\;\; \(\phi\left(x\right) = x\)};

  \draw[thick,color=black] (c3r1southwest) -- (c3r1southeast) node[midway,below] {\small Matrix Square Root};

  \draw[thick,color=black] ($(c1r1southwest) + (0,-0.7cm)$) -- ($(c1r1southeast) + (0,-0.7cm)$) node[midway,below] {\small\cref{thm:gradient_upper_bound}};
  \draw[thick,color=black] ($(c2r1southwest) + (0,-0.7cm)$) -- ($(c2r1southeast) + (0,-0.7cm)$) node[midway,below] {\small\cref{thm:gradient_upper_bound_bounded_entropy}};
  \draw[thick,color=black] ($(c3r1southwest) + (0,-0.7cm)$) -- ($(c3r1southeast) + (0,-0.7cm)$) node[midway,below] {\small\cref{thm:gradient_upper_bound}};

  \node at ($(group c2r1) + (0,2cm)$) {\ref*{grouplegend}}; 
\end{tikzpicture}
}
  \vspace{-2ex}
  \caption{
    \textbf{Evaluation of the bounds for a perfectly conditioned quadratic target.}
    The \textcolor{color3}{blue regions} are the loosenesses resulting from either using (\cref{thm:gradient_upper_bound}) or not using (\cref{thm:gradient_upper_bound_bounded_entropy}) the bounded entropy assumption (\cref{assumption:bounded_entropy}), while the \textcolor{color1}{red regions} are the remaining ``technical loosesnesses.''
    The gradient variance was estimated from \(10^3\) samples.
  }\label{fig:quadratic_add}
\end{figure}

\vspace{-1.5ex}
\subsection{Real Datasets}\label{section:additional_linearreg}
\begin{figure}[H]
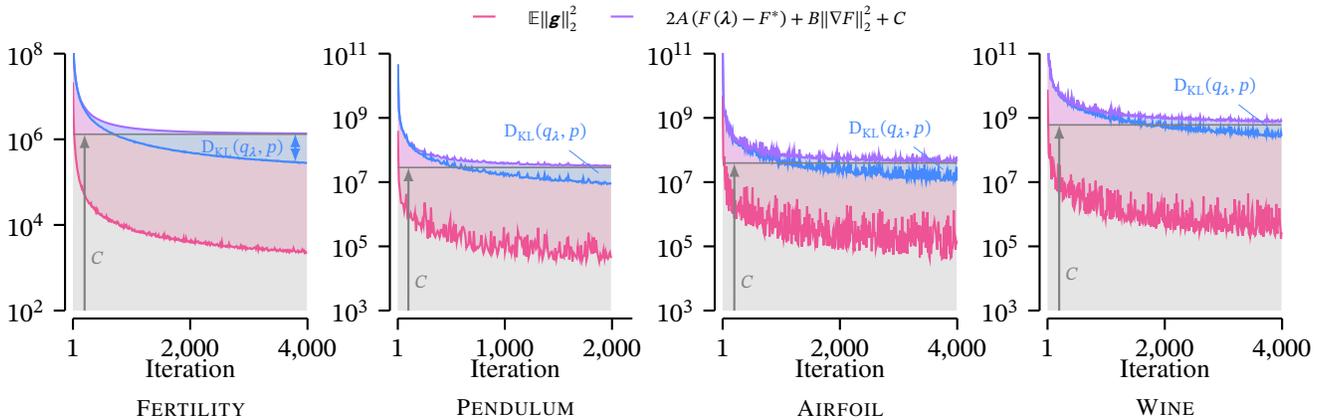

  \centering
{\hypersetup{linkbordercolor=black,linkcolor=black}
\begin{tikzpicture}
  \begin{groupplot}
    [
      group style={
        group size=4 by 1,
        horizontal sep=0.07\textwidth
      },
    ]
    \input{figures/group_fertility_squareroot_generalbound}
    \input{figures/group_pendulum_squareroot_generalbound}
    \input{figures/group_airfoil_squareroot_generalbound}
    \input{figures/group_wine_squareroot_generalbound}
  \end{groupplot}
  
  \draw[thick,color=black] ($(group c1r1) + (0cm,-3cm)$) node {\small \textsc{Fertility} };
  \draw[thick,color=black] ($(group c2r1) + (0cm,-3cm)$) node {\small \textsc{Pendulum} };
  \draw[thick,color=black] ($(group c3r1) + (0cm,-3cm)$) node {\small \textsc{Airfoil} };
  \draw[thick,color=black] ($(group c4r1) + (0cm,-3cm)$) node {\small \textsc{Wine} };

  \node at ($(group c2r1) + (2.35cm,2.2cm)$) {\ref*{grouplegend}}; 
\end{tikzpicture}
}
  \vspace{-4ex}
  \caption{
    \textbf{Evaluation of \cref{thm:gradient_upper_bound} matrix square root parameterization on linear regression datasets.}
    The gradient variance was estimated from \(4 \times 10^3\) samples.
  }\label{fig:linearreg_squareroot}
\end{figure}

We provide detailed information and additional results for the linear regression problem in~\cref{section:linearreg}.
The constants for the linear regression datasets are shown in \cref{table:datasets}, while additional results for the nonlinear Cholesky (\cref{fig:linearreg_softpluschol}), linear Cholesky (\cref{fig:linearreg_linearchol}), nonlinear mean-field (\cref{fig:linearreg_softplusmf}), and matrix square root (\cref{fig:linearreg_squareroot}) parameterizations are displayed.

{\hypersetup{linkbordercolor=black,linkcolor=black}
\begin{table*}[ht]
  \vspace{-2ex}
\caption{Properties of the Linear Regression Datasets}\label{table:datasets}
\begin{center}
  {\small
\begin{threeparttable}
\begin{tabular}{lrrrrrrrrr}
    \toprule
    \multicolumn{1}{c}{\multirow{2}{*}{\textbf{Dataset}}}
    & \multicolumn{1}{c}{\multirow{2}{*}{\(d\)}}
    & \multicolumn{1}{c}{\multirow{2}{*}{\(N\)}}
    & \multicolumn{1}{c}{\multirow{2}{*}{\(L_{\mathrm{H}}\)}}
    & \multicolumn{1}{c}{\multirow{2}{*}{\(\mu_{\mathrm{KL}}\)}}
    & \multicolumn{1}{c}{\multirow{2}{*}{\(\kappa_{\mathrm{cond.}}\)}}
    & \multicolumn{1}{c}{\multirow{2}{*}{\({\lVert \bar{\vzeta}_{\mathrm{KL}} - \bar{\vzeta}_{\mathrm{H}} \rVert}_2^2\)}}
    & \multicolumn{2}{c}{\textbf{Constants for \cref{thm:gradient_upper_bound}}} &  \\ \cmidrule{8-9}
    & & & & & & & \multicolumn{1}{c}{\(A\)} & \multicolumn{1}{c}{\(C\)}
    \\ \midrule
    \textsc{Fertility} & 9 & 100 & \(1.840 \times 10^3\) & \(5.017 \times 10^2\) &  4 & \(5.167 \times 10^{-9}\) & \(1.620 \times 10^4\) & \(1.313 \times 10^6\) \\
    \textsc{Pendulum}  & 9 & 630 & \(1.525 \times 10^4\) & \(1.897 \times 10^3\) &  8 & \(1.243 \times 10^{-10}\) & \(2.942 \times 10^5\) & \(2.858 \times 10^7\) \\
    \textsc{Airfoil}   & 5 & 1,503 & \(3.520 \times 10^4\) & \(2.909 \times 10^3\) & 12 & \(2.937 \times 10^{-10}\) & \(6.815 \times 10^5\) & \(3.936 \times 10^7\) \\
    \textsc{Wine}      & 11 & 1,599 & \(5.526 \times 10^4\) & \(1.786 \times 10^3\) & 31 & \(6.628 \times 10^{-9}\)  & \(4.787 \times 10^6\) & \(6.054 \times 10^8\) \\
    \bottomrule
\end{tabular}
\begin{tablenotes}
\item[*] \(N\) is the number of datapoints in the dataset, \(\kappa_{\mathrm{cond.}} = L_{\mathrm{H}} / \mu_{\mathrm{KL}}\) is the condition number.
\end{tablenotes}
\end{threeparttable}
  }%
\end{center}
\end{table*}
}


\begin{figure}[H]
  \vspace{-3ex}
  \centering
{\hypersetup{linkbordercolor=black,linkcolor=black}
\begin{tikzpicture}
  \begin{groupplot}
    [
      group style={
        group size=3 by 1,
        horizontal sep=0.1\textwidth
      },
    ]
    \input{figures/group_fertility_softpluschol_generalbound}
    \input{figures/group_pendulum_softpluschol_generalbound}
    \input{figures/group_wine_softpluschol_generalbound}
  \end{groupplot}
  
  \draw[thick,color=black] ($(group c1r1) + (0cm,-2.7cm)$) node {\small \textsc{Fertility} };
  \draw[thick,color=black] ($(group c2r1) + (0cm,-2.7cm)$) node {\small \textsc{Pendulum} };
  \draw[thick,color=black] ($(group c3r1) + (0cm,-2.7cm)$) node {\small \textsc{Wine} };

  \node at ($(group c2r1) + (0cm,2cm)$) {\ref*{grouplegend}}; 
\end{tikzpicture}
}
  \vspace{-2ex}
  \caption{
    \textbf{Evaluation of \cref{thm:gradient_upper_bound} with the nonlinear Cholesky (\(\phi\left(x\right) = \mathrm{softplus}\left(x\right)\)) parameterization on linear regression datasets.}
    The gradient variance was estimated from \(4 \times 10^3\) samples.
  }\label{fig:linearreg_softpluschol}
\end{figure}

\begin{figure}[H]
  \centering
{\hypersetup{linkbordercolor=black,linkcolor=black}
\begin{tikzpicture}
  \begin{groupplot}
    [
      group style={
        group size=4 by 1,
        horizontal sep=0.07\textwidth
      },
    ]
    \input{figures/group_fertility_linearchol_generalbound}
    \input{figures/group_pendulum_linearchol_generalbound}
    \input{figures/group_airfoil_linearchol_generalbound}
    \input{figures/group_wine_linearchol_generalbound}
  \end{groupplot}
  
  \draw[thick,color=black] ($(group c1r1) + (0cm,-3cm)$) node {\small \textsc{Fertility} };
  \draw[thick,color=black] ($(group c2r1) + (0cm,-3cm)$) node {\small \textsc{Pendulum} };
  \draw[thick,color=black] ($(group c3r1) + (0cm,-3cm)$) node {\small \textsc{Airfoil} };
  \draw[thick,color=black] ($(group c4r1) + (0cm,-3cm)$) node {\small \textsc{Wine} };

  \node at ($(group c2r1) + (2.35cm,2.2cm)$) {\ref*{grouplegend}}; 
\end{tikzpicture}
}
  \vspace{-4ex}
  \caption{
    \textbf{Evaluation of \cref{thm:gradient_upper_bound} with the linear Cholesky (\(\phi\left(x\right) = x\)) parameterization on linear regression datasets.}
    The gradient variance was estimated from \(4 \times 10^3\) samples.
  }\label{fig:linearreg_linearchol}
\end{figure}

\begin{figure}[H]
  \centering
{\hypersetup{linkbordercolor=black,linkcolor=black}
\begin{tikzpicture}
  \begin{groupplot}
    [
      group style={
        group size=4 by 1,
        horizontal sep=0.07\textwidth
      },
    ]
    \input{figures/group_fertility_softplusmf_generalbound}
    \input{figures/group_pendulum_softplusmf_generalbound}
    \input{figures/group_airfoil_softplusmf_generalbound}
    \input{figures/group_wine_softplusmf_generalbound}
  \end{groupplot}
  
  \draw[thick,color=black] ($(group c1r1) + (0cm,-3cm)$) node {\small \textsc{Fertility} };
  \draw[thick,color=black] ($(group c2r1) + (0cm,-3cm)$) node {\small \textsc{Pendulum} };
  \draw[thick,color=black] ($(group c3r1) + (0cm,-3cm)$) node {\small \textsc{Airfoil} };
  \draw[thick,color=black] ($(group c4r1) + (0cm,-3cm)$) node {\small \textsc{Wine} };

  \node at ($(group c2r1) + (2.35cm,2.2cm)$) {\ref*{grouplegend}}; 
\end{tikzpicture}
}
  \vspace{-4ex}
  \caption{
    \textbf{Evaluation of \cref{thm:gradient_upper_bound} with the nonlinear mean-field (\(\phi\left(x\right) = \mathrm{softplus}\left(x\right)\)) parameterization on linear regression datasets.}
    The gradient variance was estimated from \(4 \times 10^3\) samples.
  }\label{fig:linearreg_softplusmf}
\end{figure}

\clearpage
\twocolumn

\clearpage
\section{Proofs}
\subsection{External Lemmas}\label{section:external_lemmas}
\vspace{2ex}

\begin{theoremEnd}[category=common]{lemma}[\citealt{domke_provable_2019}, Lemma 9]
\label{thm:u_identities}
  Let \(\rvvu = \left(\rvu_1, \rvu_2, \ldots, \rvu_d\right)\) be a \(d\)-dimensional vector-valued random variable with zero-mean independently and identically distributed components.
  Then,
  \begin{alignat*}{2}
    &\mathbb{E}\rvvu \rvvu^{\top} &&= \left( \mathbb{E} \rvu_i^2 \right) \boldupright{I}
    \\
    &\mathbb{E}\norm{\rvvu}_2^2 &&= d \, \mathbb{E} \rvu_i^2
    \\
    &\mathbb{E} \rvvu \left( 1 + \norm{\rvvu}_2^2 \right) &&= \left( \mathbb{E} \rvu_i^3 \right) \mathbf{1}
    \\
    &\mathbb{E} \rvvu \rvvu^{\top} \rvvu \rvvu^{\top} &&= \left( \left(d - 1\right) \, {\left( \mathbb{E} \rvu_i^2 \right)}^2 + \mathbb{E}\rvu_i^4 \right) \boldupright{I}.
  \end{alignat*}
\end{theoremEnd}

\begin{theoremEnd}[category=common]{lemma}[\citealt{domke_provable_2019}, Lemma 1]
\label{thm:variational_gradient_norm_identity}
  Let \(\vt_{\vlambda}: \mathbb{R}^d \rightarrow \mathbb{R}^d\) be a location-scale reparameterization function (\cref{def:reparam}).
  Also, let \(f : \mathbb{R}^d \mapsto \mathbb{R} \) be some differentiable function.
  Then,
  \begin{alignat*}{2}
    &\norm{\nabla_{\vlambda} f\left( \vt_{\vlambda}\left(\vu\right) \right) }_2^2 \\
    &\;= 
    \norm{\nabla f\left( \vt_{\vlambda}\left(\vu\right) \right) }_2^2 \left(1 + \norm{\vu}_2^2\right).
  \end{alignat*}
\end{theoremEnd}

\begin{theoremEnd}[category=common]{lemma}[\citealt{domke_provable_2019}, Lemma 1]
\label{thm:reparam_u_identity}
  Let \(\vt_{\vlambda}: \mathbb{R}^d \rightarrow \mathbb{R}^d\) be a location-scale reparameterizaiton function (\cref{def:reparam}).
  Also, let \(\vz \in \mathbb{R}^d\) be some vector and \(\rvvu \sim \varphi\) satisfy~\cref{assumption:symmetric_standard}.
  Then,
  \begin{alignat*}{2}
    \mathbb{E} \norm{\vt_{\vlambda}\left(\rvvu\right) - \vz}_2^2 \left(1 + \norm{\rvvu}_2^2\right)
    =
    \left(d+1\right) \norm{\vm - \vz}_2^2 + \left(d + \kappa\right) \norm{\mC}^2_{\mathrm{F}}.
  \end{alignat*}
\end{theoremEnd}


\printProofs[upperboundlemma]

\clearpage
\subsection{Proof of Key Lemmas}
\subsubsection{Proof of \cref*{thm:general_variational_gradient_norm_identity}}
\vspace{2ex}
\printProofs[upperboundkeylemmagradientnormidentity]
\newpage
\subsubsection{Proof of \cref*{thm:general_variational_gradient_norm_bound}}
\vspace{2ex}
\printProofs[upperboundkeylemmagradientnormbound]
\newpage
\subsubsection{Proof of \cref*{thm:meanfield_u_identity}}
\vspace{2ex}
\printProofs[upperboundkeylemmameanfield]
\newpage
\subsubsection{Proof of \cref*{thm:gradient_variance_general_upper_bound}}
\vspace{2ex}
\printProofs[upperboundkeylemmavariancegeneral]

\clearpage
\subsection{Proof of Theorems}
\subsubsection{Proof of \cref*{thm:gradient_upper_bound}}
\vspace{2ex}
\printProofs[upperboundtheorem]
\newpage
\subsubsection{Proof of \cref*{thm:gradient_upper_bound_kl}}
\vspace{2ex}
\printProofs[upperboundtheoremklform]
\newpage
\subsubsection{Proof of \cref*{thm:gradient_upper_bound_bounded_entropy}}
\vspace{2ex}
\printProofs[upperboundtheoremboundedentropy]

\newpage
\subsubsection{Proof of \cref*{thm:gradient_lower_bound}}
\vspace{2ex}
\printProofs[lowerboundtheorem]





\end{document}